\documentclass{article}

\usepackage{amsmath,amsfonts,bm}
\def\secref#1{section~\ref{#1}}
\def\eqref#1{equation~\ref{#1}}
\def\1{\bm{1}}

\DeclareMathOperator*{\argmin}{arg\,min}

\usepackage[preprint]{neurips_2025}

\usepackage[utf8]{inputenc} % allow utf-8 input
\usepackage[T1]{fontenc}    % use 8-bit T1 fonts
\usepackage[dvipsnames]{xcolor}         % colors
\definecolor{navy}{rgb}{0.0, 0.0, 0.5}
\usepackage[backref]{hyperref}
\hypersetup{
    colorlinks=true,
    linkcolor=CadetBlue,
    filecolor=magenta,      
    urlcolor=CadetBlue,
    citecolor=CadetBlue,
    pdftitle={Hessian Learning},
}
\usepackage{url}            % simple URL typesetting
\usepackage{booktabs}       % professional-quality tables
\usepackage{nicefrac}       % compact symbols for 1/2, etc.
\usepackage{microtype}      % microtypography
\usepackage{placeins}
\usepackage{amssymb}
\usepackage{mathtools}
\usepackage{amsthm}
\usepackage{chemformula}
\usepackage{makecell}
\usepackage{subcaption}
\usepackage{tcolorbox}
\usepackage{float}
\usepackage{todonotes}
\theoremstyle{definition}

\usepackage{accents}

\usepackage{multirow}

\newcommand{\equiformer}{EquiformerV2}

\usepackage{fancyhdr}
\title{HIP:\\ Hessian Interatomic Potentials\\ Without Derivatives}

\author{%
Andreas Burger\textsuperscript{1,2}\thanks{\{andreas.burger\}\{luca.thiede\}@mail.utoronto.ca}\;
\ Luca Thiede\textsuperscript{1,2}\footnotemark[1]\\[1ex]
\textbf{Nikolaj Rønne}\textsuperscript{3,4}\;
\textbf{Varinia Bernales}\textsuperscript{1,5}\;
\textbf{Nandita Vijaykumar}\textsuperscript{1}\\[1ex]
\textbf{Tejs Vegge}\textsuperscript{3,4}\;
\textbf{Arghya Bhowmik}\textsuperscript{3,4}\;
\textbf{Alán Aspuru-Guzik}\textsuperscript{1,2,5,6,7}\thanks{aspuru@nvidia.com}\\[1ex]
\textsuperscript{1}University of Toronto \\
\textsuperscript{2}Vector Institute for Artificial Intelligence \\
\textsuperscript{3}Technical University of Denmark \\
\textsuperscript{4}CAPeX Pioneer Center for Accelerating P2X Materials Discovery \\
\textsuperscript{5}Acceleration Consortium \\
\textsuperscript{6}Canadian Institute for Advanced Research (CIFAR) \\
\textsuperscript{7}NVIDIA \\[1ex]
}
\begin{document}
\maketitle
\begin{abstract}
Molecular Hessians, the second derivatives of the potential energy, are fundamental to many workflows in computational chemistry. Usually, accurate Hessians are computationally expensive to calculate and scale poorly with system size, whether computed using quantum chemistry methods or machine-learning interatomic potentials (MLIPs). In this work, we introduce Hessian interatomic potentials (HIPs), a deep learning model that directly predicts Hessians without relying on automatic differentiation or finite differences. To do so, we construct SE(3)-equivariant, symmetric Hessians from irreducible representation (irrep) features up to degree $l$=2, computed by a graph neural network. HIP Hessians are one to two orders of magnitude faster, more accurate, more memory efficient, easier to train, and exhibit more favourable scaling with system size. We validate our predictions across a wide range of downstream tasks, demonstrating consistently superior performance in transition state search, geometry optimization, zero-point energy corrections, and vibrational analysis. We open-source the HIP code and model weights.
\end{abstract}

\section{Introduction}
Molecular simulation is a cornerstone of material discovery and molecular design. While the prediction of energy and forces is often sufficient for structural dynamics, advanced computational chemistry workflows often depend on the molecular Hessian\citep{qu2016,huang2018}. By describing the local curvature of the potential energy surface (PES), the Hessian is foundational to many computational workflows. Second-order geometry optimization relies on the curvature information to rapidly and reliably converge to equilibrium structures\cite{schlegel2011}. Transition state (TS) searches and extrema classification depend on the Hessian to correctly identify saddle points and reaction barriers, which are essential for parameterizing reaction rates and microkinetic models for longer time and length scale simulations\cite{andersen2019}. Furthermore, mode-following transition state search methods directly leverage the curvature information to locate the transition states\cite{henkelman1999}. Finally, vibrational analysis uses Hessian eigenvalues to compute zero-point energy (ZPE) corrections and thermodynamic free energies bridging the gap from static molecules to non-zero temperature observables\cite{grimme2012}. 

For a molecule with $N$ atoms, the Hessian is a $3N \times 3N$ matrix, where each entry requires a mixed second-order derivative of the electronic energy with respect to two nuclear coordinates. Despite their broad utility, Hessian calculations represent a substantial computational bottleneck.
Traditional approaches to obtain the Hessian rely either on analytic second derivatives or on numerical differentiation of gradients, both of which are very expensive, and analytical Hessians in particular are often so complicated to implement that they are not widely supported for all methods \citep{jorgensen1983,handy1984,komornicki1993}.

Most computational chemistry calculations are carried out with Kohn-Sham density functional theory \citep{kohn1965} (DFT), due to its balance of computational cost and accuracy \citep{bursch2022}. However, the $O(N^4)$ scaling of hybrid DFT with system size remains expensive, which has spurred the development of machine-learning interatomic potentials (MLIPs). MLIPs promise to deliver near-DFT accuracy at a fraction of the cost while incorporating symmetries, thereby drastically reducing the need for extensive training data \citep{schutt2018,thomas2018,gasteiger2021,schutt2021,batatia2022,batzner2022}.
While recent work demonstrates that automatic differentiation (AD) can successfully recover the Hessian \citep{yang2024, yuan2024, gonnheimer2025, rodriguez2025, williams2025, cui2025horm}, this approach faces two substantial limitations. High accuracy of the energy and forces does not guarantee comparably accurate second derivatives. Much like training on force labels is needed to achieve accurate force predictions \cite{christensen2020forces}, explicit Hessian training labels are required for reliable AD Hessians \citep{zhao2025, cui2025horm}.
Second, these AD Hessians introduce high computational costs, with an extra factor of $\sim{}3N$ during training and inference \citep{yuan2024,gonnheimer2025,rodriguez2025,williams2025,cui2025horm}. 
Rodriguez et al. ~\cite{rodriguez2025} in particular showed that including the Hessian in the loss improves data efficiency and extrapolation of energy and force predictions but is limited by approximately 25 times longer training times, which make AD Hessians prohibitive to train. The same principle extends to learned exchange-correlation functionals, where supervising energy derivatives substantially improves accuracy \citep{eberhard2026derivative}.
MLIPs can also be used to obtain Hessians with finite differences \citep{gonnheimer2025, fu2025}, an approach which requires less memory but $6N$ forward passes, and also requires specialized data collection schemes to be accurate \cite{wander2025accessing}.\\
In this work, we introduce \textbf{HIP}: \textbf{H}essian \textbf{I}nteratomic \textbf{P}otentials. HIPs predict the Hessian directly, to eliminate the need for finite differences, coupled-perturbed equation solvers, or automatic differentiation.
By building on SE(3)-equivariant neural networks, we demonstrate how HIP can predict Hessians at a fraction of the computational cost of traditional methods. 
A key observation is that one can construct the Hessian from spherical-harmonic features of the atom nodes and the messages between atoms, satisfying equivariance and symmetry constraints by design. 
In practice, HIP is not only faster but also more accurate than AD, enabling a variety of critical tasks in molecular simulations.
Our contributions are the following: 
\begin{enumerate}
    \item We present Hessian Interatomic Potentials (HIP), a method to construct SE(3)-equivariant, symmetric Hessians \emph{directly}, without finite differences, coupled-perturbed solvers, or automatic differentiation, using an equivariant neural network backbone.
    %\item We introduce a loss function that targets the part of the Hessian subspace particularly important for molecular optimization problems.
    \item We show that downstream results can be improved further by using a loss function that targets the relevant Hessian subspace
    \item We show that, compared to AD, the predicted Hessians are 10-100x faster, while also requiring 2-3x less peak memory on even small molecules with 5-30 atoms, with better saling.
    \item We show that the predicted Hessians are more accurate on benchmark datasets, with 2x lower Hessian MAE, 3.5x lower eigenvalue MAE, and 1.5x higher eigenvector cosine similarity, and significantly better generalization to larger systems. 
    \item Finally, we validate the practical utility of our predicted Hessians in several downstream applications, such as (i) zero-point energy corrections, (ii) second-order geometry optimization, (iii) transition state search, and (iv) frequency analysis for extrema classification. 
\end{enumerate}

\begin{figure}
    \centering
    \includegraphics[width=0.99\linewidth]{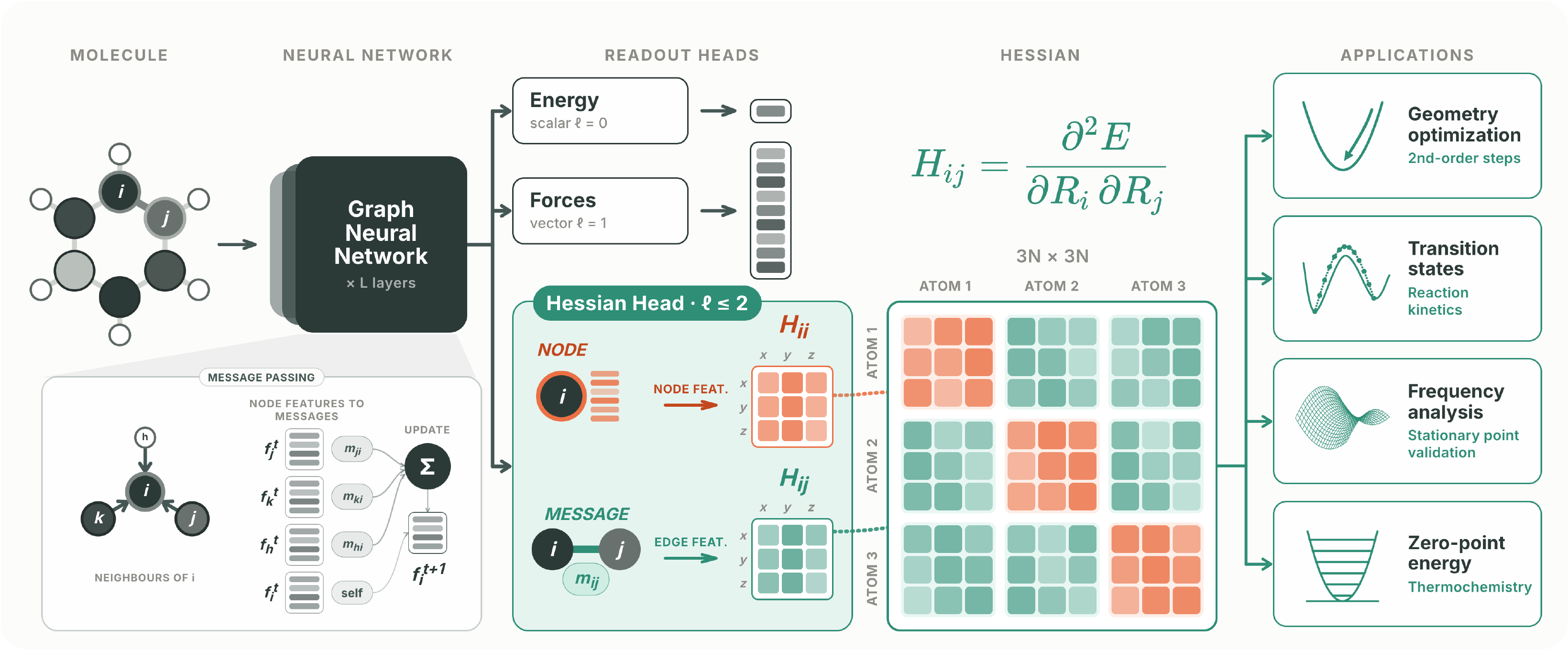}
    \caption{
    \textbf{HIP architecture.}
    A conventional MLIP (grey) maps a molecular geometry to its energy and forces.
    HIP adds an equivariant Hessian readout that combines node contributions (orange) and pairwise message contributions (green) to predict each Hessian block $H_{ij}=\partial^2 E/\partial\bm{R}_i\partial\bm{R}_j$ directly.
    Conventional MLIPs instead construct the Hessian by automatic differentiation (AD) of the predicted energy or forces.
    The predicted Hessian supports geometry optimization, transition-state searches, frequency analysis and zero-point-energy calculations.
    HIP can use both direct force prediction and conservative forces.
    }
    \label{fig:visual_abstract}
\end{figure}

\section{Background}
\paragraph{MLIPs and equivariant neural networks}
Early works on machine learning interatomic potentials used local atomic environment descriptors in combination with linear regression \citep{thompson2015,shapeev2016}, Gaussian processes \citep{bartok2010}, and simple neural networks \citep{behler2007}.
SchNet \citep{schutt2018} was the first work to introduce \textit{rotation-invariant graph neural networks} to predict molecular properties like energies and forces. Next-generation invariant architectures include ViSNet \citep{wang2022} and QuinNet \citep{wang2024}.
In parallel, \textit{rotation-equivariant} architectures were developed, such as Cormorant \citep{anderson2019}, DimeNet \citep{gasteiger2020}, GemNet \citep{gasteiger2021}, PaiNN \citep{schutt2021}, NequIP \citep{batzner2022}, SphereNet \citep{liu2022}, MACE \citep{batatia2022} and the Equiformer family \citep{liao2023, liao2023equiformerv2}. 
\emph{Graph neural networks} represent a molecule as a graph $\mathcal{G}=(\mathcal{V},\mathcal{E})$ in 3D Euclidean space $\mathbb{R}^3$, with nodes/atoms $I\in\mathcal{V}$ at positions $\mathbf{r}_I\in\mathbb{R}^3$ and edges $(I,J)\in\mathcal{E}$ defined by a cutoff radius. At layer $t$, each node carries a feature $\mathbf{h}_I^{(t)}$ that is a direct sum of irreducible $\mathrm{SO}(3)$ representations (irreps): $\mathbf{h}_I^{(t)}=\bigoplus_{l=0}^{l_{\max}}\mathbf{h}_{I}^{(t,l)}$, where $\mathbf{h}_{I}^{(t,l)}=\{\mathbf{h}^{(t)}_{I,l,m}\}_{m=-l}^{l}$ has $2l+1$ components. To update the node's features, the model sends messages between each connected node, which are then aggregated in a permutation invariant fashion (usually the sum, or attention-weighted-sum operation) and then added to the receiving node.
In $\mathrm{SO}(3)$-equivariant GNNs, node features transform under a global rotation $\mathbf{Q}$ via Wigner $\mathbf{D}$-matrices:
\begin{align}
\mathbf{h}^{(t)}_{i,l,m}\!\left(\mathbf{Q}\{\mathbf{r}_k\}_{k=1}^N\right)
= \sum_{m'=-l}^{l} \mathbf{D}^{(l)}_{m m'}(\mathbf{Q})\,\mathbf{h}^{(t)}_{i,l,m'}\!\left(\{\mathbf{r}_k\}_{k=1}^N\right),
\label{eq:so3-transform}
\end{align}
Translation equivariance is enforced by building messages from relative displacements $\mathbf{r}_{I,J}=\mathbf{r}_J-\mathbf{r}_I$. If we want O(3) instead of SO(3) equivariance, each order-$l$ feature has an extra parity label that determines the transformation under a global coordinate inversion.
\\
% A typical equivariant message-passing update couples neighbor features with equivariant edge features via Clebsch-Gordan (CG) tensor products:
% \begin{align}
% \mathbf{h}^{(t+1)}_{i,l}
% = \sum_{j\in\mathcal{N}(i)} \sum_{l_1,l_2}
% W^{(t)}_{l_1,l_2\!\to l}\,
% \big(\mathbf{h}^{(t)}_{j,l_1} \otimes \mathbf{e}^{(t)}_{ij,l_2}\big)\Big|_{l},
% \qquad
% \mathbf{e}^{(t)}_{ij,l_2,m}
% = \phi^{(t)}_{l_2}(r_{ij})\,Y_{l_2}^{m}(\widehat{\mathbf{r}}_{ij}),
% \label{eq:update}
% \end{align}
% where $r_{ij}=\|\mathbf{r}_{ij}\|$, $\widehat{\mathbf{r}}_{ij}=\mathbf{r}_{ij}/r_{ij}$, $Y_{l}^{m}$ are spherical harmonics, $\phi^{(t)}_{l}$ are learnable radial functions (e.g., RBFs passed through an MLP), $\otimes$ denotes the CG product, and $(\cdot)|_{l}$ projects onto the order-$l$ irrep. Scalar ($l{=}0$) biases and nonlinearities are inserted in an irrep-wise manner to preserve equivariance.
Two irreps $\mathbf{p}_{l_1}$ and $\mathbf{g}_{l_2}$ interact using the Clebsch-Gordan tensor product \citep{thomas2018}
\begin{align}
    \mathbf{h}_{l_3,m_3} = \left( \mathbf{p}_{l_1,m_1} \otimes_{l_1, l_2}^{l_3} \mathbf{g}_{l_2,m_2} \right)_{m_3} = \sum_{m_1 = -l_1}^{l_1} \sum_{m_2 = -l_2}^{l_2} \mathbf{C}_{(l_1,m_1),(l_2,m_2)}^{(l_3,m_3)} \mathbf{p}_{l_1,m_1} \mathbf{g}_{l_2,m_2} \label{eq:tp}
\end{align}
A message from source atom $J$ to target atom $I$ is then constructed as
\begin{align}
    \mathbf{v}_{I,J, l_3} = \mathbf{v}_{l_3}(\mathbf{h}_I, \mathbf{h}_J, \mathbf{r}_{I,J}) = \sum_{l_1, l_2} w_{l_1, l_2, l_3}(||\mathbf{r}_{I,J}||) \left( \mathbf{f}_{l_1}(\mathbf{h}_{I},\mathbf{h}_{J}) \otimes_{l_1, l_2}^{l_3} \mathbf{Y}_{l_2}\left(\frac{\mathbf{r}_{I,J}}{||\mathbf{r}_{I,J}||}\right) \right) \label{eq:mp}
\end{align}
where $\mathbf{Y}_{l_2}$ is the $2l_2+1$ dimensional vector of the spherical harmonics of degree $l_2$, $w_{l_1, l_2, l_3}: \mathbb{R} \rightarrow \mathbb{R}$ is a learned weighting function, and $\mathbf{f}_{l_1}$ a simple function that takes both node features in, for example concatenation in the case of \equiformer{}.
A more efficient variation of \eqref{eq:mp} relying on the reduction of the $SO(3)$ tensor product to the $SO(2)$ tensor products was proposed by eSCN \citep{eSCN2023} and adopted by subsequent works, such as \equiformer{}.
Finally, the messages are pooled to update the node features, which can be done, for example, via a sum or using graph attention \citep{liao2023equiformerv2}.
\\
After several layers of message passing, task-dependent readout heads map the node features to the desired targets while respecting symmetry. For example, energies $E$ are $\mathrm{SO}(3)$-invariant per-graph scalars. The energy readout head therefore uses a global pooling and a reduction operation to output a $l{=}0$ feature. Forces $\mathbf{F}$ are $\mathrm{SO}(3)$ equivariant per-atom vectors ($l=1$), and can be outputted either by a direct force readout head, or as the derivative of the energy $\mathbf{F} = -\nabla_\mathbf{R} E$. As we will see in \secref{sec:method}, Hessians are $\mathrm{SO}(3)$-equivariant per-graph matrices composed of per-atom-pair $3\times3$ equivariant sub-matrices. 
% We will explain how to construct such a readout head for Hessians below.
% 

% \subsection{Equivariant matrix prediction}
\paragraph{Equivariant matrix prediction}
There is a rich connection between the HIP Hessian prediction in this work and other rotation-equivariant matrices of interest, like Hamiltonian prediction \cite{kaniselvan2025helm}. For example, PhisNet \cite{unke2021se}, QHNet \cite{yu2023efficient}, HELM \cite{kaniselvan2025helm}, and WANet \cite{li2025enhancing} predict the Fock matrix, and Graph2Mat \cite{febrer2025graph2mat} targets the density matrix.

\subsection{Traditional methods for calculating Hessians}
\paragraph{Hessians from DFT}
The gold standard for computing nuclear Hessians is to derive them analytically. In Kohn-Sham DFT, this is achieved using the Coupled Perturbed Kohn-Sham (CPKS) method \citep{deglmannEfficientImplementationSecond2002}, which is employed to calculate the ground truth in the dataset and in the workflows of our experiments. If available, analytical Hessians provide the most efficient ab-initio way to obtain Hessians, but they still scale as $O(N^5)$. See Appendix \ref{sec:cpks} for details.
% \paragraph{Finite differences of analytic gradients.}
A major downside of the CPKS approach to Hessians is its highly complex implementation and need for exchange-correlation-functional-specific derivations, which makes it not always available. An alternative is to use a finite-difference scheme, requiring only energies or forces. Finite differences also scale $O(N^5)$ for hybrid DFT, although with more numerical noise and a much larger computational prefactor. We provide more details in \ref{sec:finite_difference}.
\paragraph{Hessians from MLIPs}
MLIPs have gained widespread popularity for their ability to accurately approximate molecular energies and forces. The forces can either be directly predicted \citep{gasteiger2021, liao2023equiformerv2}, or calculated as the derivative of the MLIP energy using automatic differentiation. %, with a cost of about 2-3 times that of an energy forward pass. 
Previous work has shown that the Hessian can be calculated using AD to some success \citep{yang2024, yuan2024, gonnheimer2025, rodriguez2025, williams2025, cui2025horm}. 
Although AD Hessians are much cheaper to calculate compared to DFT, they still have some downsides.
%A natural question then is to ask if we can also calculate the Hessian cheaply by simply differentiating the energy twice. In principle, the answer is yes, but with some nuances:
First, accurately modeling the energy does not guarantee low errors on Hessians \citep{yuan2024}, but requires dedicated training \citep{zhao2025, cui2025horm}. Second, compared to energies and forces, Hessians are expensive: 
%Automatic differentiation calculates the Hessian via Hessian-Vector-Products (HVPs). 
%Given a probing vector $\mathbf{v}_i = \mathbf{e}_i, i=0,...,3N-1$, AD allows us to calculate the $i$-th column of the Hessian with $\mathbf{H}\mathbf{v}_i$. 
Given a probing vector $\mathbf{v}$, AD only lets us calculate $\mathbf{H}\mathbf{v}$. 
To calculate the full $3N\times 3N$ Hessian, we need $3N$ Hessian vector products (HVPs) with the unit vectors $\mathbf{v} = \mathbf{e}_i, i=0,...,3N-1$, each yielding one column of the Hessian. 
Since the energy calculation with an MLIP is usually $O(N)$, the total cost of the Hessian calculation is $O(N^2)$. 
While the HVPs can theoretically be parallelized, the $O(N^2)$ memory footprint requires even medium-sized systems to be computed sequentially.
% Alternatively, we can also get Hessians with finite differences \citep{gonnheimer2025, fu2025}.
\cite{rodriguez2025} in particular showed that including the Hessian in the loss leads to better extrapolation of energy and force predictions, less data for a given accuracy, and improved stability in MD. This comes at the cost of approximately 25 times longer training times, which makes AD Hessians prohibitive to train.

\subsection{Molecular optimization with Hessians}

\paragraph{Rational function optimization (RFO)}
Geometry optimization locates minima $\argmin_\mathbf{R} E(\mathbf{R})$ of the potential energy surface (PES) starting from non-equilibrium molecular geometries. It underpins most workflows from global searches to high-throughput screening. 
% There is a rich set of optimizers that have been developed, some of which we review in \ref{sec:optimizerliterature}.
If Hessians are available, second-order methods are favored due to their fast convergence guarantees \citep{nocedalNumericalOptimization2006, doikovSecondOrderOptimizationLazy2023}. 
% \\
RFO \citep{simons1983,banerjee1985} is a commonly used second-order optimizer (it makes use of Hessian information to accelerate convergence) for molecular geometries \citep{sella2022}. 
% RFO starts with a [2/2]-Pad\'e-expansion of the energy:
% \begin{align}
%     E(\mathbf{R_t} + \Delta \mathbf{x}) - E(\mathbf{R_t)} \approx \frac{\mathbf{g}^\top \Delta \mathbf{x} + \frac{1}{2} \Delta \mathbf{x}^\top \mathbf{H} \Delta \mathbf{x}}{1 + |\Delta\mathbf{x}|^2 }
% \end{align}
% The extrema $\Delta x'$ of this surrogate are given by the solution of the generalized eigenvalue problem:
% \begin{align}
% \begin{bmatrix}
%     \mathbf{H} & \mathbf{g} \\
%     \mathbf{g}^\top & 0
% \end{bmatrix}
% \begin{bmatrix}
%     \Delta \mathbf{x}' \\
%     1
% \end{bmatrix}
% =
% \lambda
% \begin{bmatrix}
%     I & 0 \\
%     0 & 1
% \end{bmatrix}
% \begin{bmatrix}
%     \Delta \mathbf{x}' \\
%     1
% \end{bmatrix}. \label{eq:augmented_eig}
% \end{align}
An attractive property of RFO is that it can be used both for minimizing $E(\mathbf{R})$ as well as for finding saddle points, and it is robust to indefinite Hessians \citep{banerjee1985}.
% If we pick the first eigenvector of \eqref{eq:augmented_eig}, we get an update pointing to the minimum; if we select the second eigenvector, we get a transition state update. 
% In contrast to Newton-Raphson optimization, RFO is robust to indefinite Hessians \citep{banerjee1985}. 
% In practice one often uses restricted step RFO (RS-RFO), which adds a trust region to prevent unphysically large steps. 
% For transition point search we are using restricted step partitioned RFO (RS-P-RFO), which treats the subspace with negative and positive eigenvalues of the Hessian separately.
% RFO requires a Hessian at each step. 
As computing the Hessian is usually too expensive, a common practice is to maintain an approximate Hessian using the BFGS quasi-Newton scheme. We provide more details in \ref{sec:bfgs}.

\paragraph{Zero-point energy (ZPE)}
Zero-point corrections account for the quantum mechanical vibrational energy that molecules have at absolute zero temperature. To calculate the ZPE, one relaxes a geometry to an extremum, computes the Hessian, and sums the frequencies
% \begin{align}
$
    ZPE = \frac{\hbar}{2} \sum_i \sqrt{\tilde{\lambda}_i} % \text{ZPE}
$.
% \end{align}
Here, $\tilde{\lambda}_i$ are eigenvalues of the mass-weighted Hessian $\mathbf{\tilde{H}}_{ij} = \mathbf{H}_{ij}/\sqrt{m_i m_j}$, and $\hbar$ is Planck's reduced constant.
% The ZPE is added to the total energy in many calculations that involve thermodynamic estimates.
One is usually interested in the relative ZPE between reactant and product states: $\Delta ZPE = ZPE(\mathbf{R}_P) - ZPE(\mathbf{R}_R)$. The relative ZPE is relevant for reaction thermochemistry, as $\Delta ZPE$ enters as a correction to the reaction free energy $\Delta G$ and therefore the equilibrium constant, which relates forward and backward rates via detailed balance.
% For reaction rates (Eyring/TST), the relevant quantity is the ZPE correction to the barrier: ΔZPE‡ = ZPE(TS) − ZPE(reactant). This shifts ΔG‡ and thus k ∝ exp(−ΔG‡/kBT).

\paragraph{Transition states}
%Why are they useful, details
%Transition state search methods, growing string method, gentlest ascend dynamics, notch elastic band
Transition states (TS) are first-order saddle points on the PES. They represent the maximum barrier along the minimum energy path (MEP) between two minima. Identifying transition states is crucial for understanding reaction mechanisms and mapping reaction networks. A first-order saddle point is characterized by having exactly one negative eigenvalue of the Hessian. Over the years, various computational methods have been developed to describe MEPs and locate transition states. We review methods for transition state search in \ref{sec:tssearchliterature}.

\paragraph{Frequency analysis for extrema classification}
To carry out frequency analysis, one must first remove the five or six redundant translational and rotational degrees of freedom, depending on whether the molecule is linear. This is done by mass-weighting the Hessian and performing an Eckart projection \citep{louck1976}. Then, the projected matrix is decomposed into its eigenvalues, with all positive eigenvalues signaling a minimum, and exactly $N$ negative eigenvalues signaling the presence of an order-$N$ transition state.
%The final step in a transition-state (TS) search or minima optimization is to confirm convergence via frequency analysis. As mentioned above, a transition state is identified by the Hessian having exactly one negative eigenvalue (or equivalently, one imaginary frequency), whereas a minimum contains no imaginary frequencies. 

\section{HIP Hessian Prediction\label{sec:method}}
Any equivariant neural network \citep{duvalHitchhikerGuideGeometric2024a} with features of at least $l=2$ can be equipped with a Hessian readout head. In this work, we use EquiformerV2 with four transformer layers as the backbone \citep{liao2023equiformerv2}. Each layer contains a layer norm, message passing with graph attention, another layer norm, and a feed-forward layer. Rather than differentiating the scalar energy output, HIP adds a readout that constructs the full Hessian directly from the equivariant node features and interatomic messages. This makes the Hessian prediction parallel to the ordinary forward pass.
%Energy and forces are both predicted directly. %For details, see \citep{liao2023equiformerv2}.

\subsection{Hessian Symmetry Requirements}
Hessians are symmetric, real-valued, Cartesian tensors. This means, under rotation of the coordinate system with rotation matrix $\mathbf{Q}$, each Hessian sub-block $\mathbf{H}_{I,J}$ transforms as 
\begin{align}
    \mathbf{H}_{I,J} \xrightarrow[]{\mathbf{Q}} \mathbf{Q} \mathbf{H}_{I,J} \mathbf{Q}^\top \label{eq:hessian_symmetry}
\end{align}
% \begin{align}
%     \mathbf{H} \xrightarrow[]{\mathbf{Q}} (\mathbf{I}_{3N}\otimes\mathbf{Q}) \mathbf{H} (\mathbf{I}_{3N} \otimes\mathbf{Q})^\top \label{eq:hessian_symmetry}
% \end{align}
% where $\mathbf{I}_{3N}$ is the $3N\times 3N$ dimensional identity matrix. 
% In particular, each sub-block corresponding to atom $I$ and $J$ transforms individually as 
% \begin{align}
%     \mathbf{H}_{I,J} \xrightarrow[]{\mathbf{Q}} \mathbf{Q} \mathbf{H}_{I,J}\mathbf{Q}^\top \label{eq:hessian_symmetry}
% \end{align}
Any physically meaningful Hessian has to satisfy this rotation symmetry, as well as the symmetry of the upper and lower triangular $\mathbf{H}=\mathbf{H}^\top$. We elaborate on the symmetry of Hessians in \ref{sec:hessian_properties}.

\subsection{Hessian Prediction Head}
% A Hessian prediction head can be built on top of any message-passing, equivariant MLIP architecture \citep{duvalHitchhikerGuideGeometric2024a}, but in this paper we are using EquiformerV2 \citep{liao2023equiformerv2}.
Starting from $T$ layers of message passing in the backbone, we use the atom features $\mathbf{h}^{(T)}_{I,m,c}$ as input to a Hessian readout module. We first add a Hessian-readout-specific EquiformerV2 layer to improve expressivity. The Hessian is decomposed into $3\times3$ blocks $\mathbf{H}_{I,J}$, with diagonal blocks describing self-interactions and off-diagonal blocks describing pair interactions. Node features are used for the self-interaction terms, while messages between atoms $I$ and $J$ are retained as pair features rather than being aggregated by attention:
\begin{align}
    \mathbf{h}_{I,J,l,m,c} 
    = \mathbf{v}_{I,J,l,m,c} 
    = \mathbf{v} \big(\mathbf{h}^{(T)}_I, \mathbf{h}^{(T)}_J, \mathbf{r}_{I,J} \big)
    % = \sum_{l_1, l_2} w_{l_1, l_2, l_3}(||\mathbf{r}_{I,J}||) \left( \mathbf{f}_{l_1}(\mathbf{h}_{I},\mathbf{h}_{J}) \otimes_{l_1, l_2}^{l_3} \mathbf{Y}_{l_2}\left(\frac{\mathbf{r}_{I,J}}{||\mathbf{r}_{I,J}||}\right) \right) 
\end{align}
By reusing the message-passing machinery with an interaction cutoff, the readout only constructs blocks for atom pairs within that cutoff. Its cost is initially $O(N^2)$ for dense small systems but becomes $O(N)$ for larger systems at fixed density, where the number of neighbors per atom remains bounded. The Hessian interaction cutoff can be set independently of the backbone radius to represent the locality of quantum-mechanical interactions.
This sparsity-with-distance assumption aligns with the sparsity of Hessians due to the locality of electron interactions, which is common in quantum chemistry and can be mathematically rigorously justified \citep{kussmann2015reduced}.
% As we are reusing the message passing machinery, it becomes trivial to use a cutoff radius to efficiently predict Hessians sparsely, leading to $O(N)$ scaling in memory and compute for larger systems. 
% Assuming that Hessians are sparse due to locality of electron interactions is common in quantum chemistry and mathematically well justified \citep{kussmann2015reduced}. 
% For simplicity we use the same cutoff radius for the Hessian as for the backbone, as it is already large enough \ref{tab:hyperparameters}.
% We inherent the cutoff radius of 12 Angstrom from the base model. 
After message passing we have to transform the atom pair features $\mathbf{h}_{I,J,l,m,c}$ into the corresponding Hessian sub-blocks $\mathbf{H}_{I,J}$.
First, we project the pair feature irreps down to a single $\{\mathbf{\tilde{h}}_{I,J,l,m}\}_{l\in\{0,1,2\}}$ irreps feature ($\texttt{1x0e + 1x1e + 1x2e}$ in \texttt{e3nn} notation \citep{geiger2022e3nn}) using a linear layer $\mathbf{W}_{l,c}$:
% \begin{align}
$
\mathbf{\tilde{h}}_{I,J,l,m} = \mathbf{W}_{l,c}\mathbf{h}_{I,J,l,m,c}
$
% \end{align} 
We then expand $\mathbf{\tilde{h}}_{I,J,l,m}$  to an intermediate Hessian sub-block $\mathbf{H}'_{I,J}$ using the tensor product expansion \citep{unke2021se}:
\begin{align}
\mathbf{H}'_{I,J,m_1,m_2} &= \sum_{l,m}\mathbf{C}^{l,m}_{l_1=1,m_1,l_2=1,m_2}\mathbf{\tilde{h}}_{I,J,l,m} \label{eq:tensor_expansion}
\end{align}
where $\mathbf{C}^{l,m}_{l_1,m_1,l_2,m_2}$ are the Clebsch--Gordan coefficients ensuring that the relation in \eqref{eq:hessian_symmetry} is enforced \citep{unke2021se}. Physically, this is a simple change of basis from the coupled to the uncoupled angular momentum basis \citep{edmonds1996angular}.
$\mathbf{H}'_{I,J,m_1,m_2}$ is a $3\times3$ block, since $m_1$ and $m_2$ run from $-l_1=-l_2=-1$ to $l_1=l_2=1$. As $\mathbf{C}^{l,m}_{l_1=1,m_1,l_2=1,m_2} = 0$ for all $l>2$, $\mathbf{h}_{I, J,l,m,c}$ only needs to contain irreps up to $l=2$ to build the $3\times3$ Hessian sub-blocks. Intuitively, \eqref{eq:tensor_expansion} is the inverse of the irreducible tensor decomposition, so it assembles a spherical tensor back from its irreducible components. Conversely, this also means we need all irreps up to $l=2$ as only including irreps up to $l=1$ would not be sufficient to express the $3\times3$ tensors that decompose into $l=2$ irreps.
% \\
%Since two irreps of degree $l=1$ can only couple to output irreps of degree up to $l=2$, $\mathbf{h}_{I, J,l,m,c}$ only needs to contain irreps up to $l=2$ to build the $3\times3$ Hessian sub-blocks.
% Limiting to $l=1$ limits to low-rank/...
% Any higher $l>2$ do not contribute to $l=1$ features.
% We now have to transform the atom pair features $\mathbf{h}_{I,J,l,m,c}$ into the corresponding Hessian sub-blocks $\mathbf{H}_{I,J}$ using the tensor product expansion \citep{unke2021se}:
% \begin{align}
% \mathbf{\tilde{h}}_{I,J,l,m} &= \mathbf{W}_{c}\mathbf{h}_{I,J,l,m,c} \\
% \mathbf{H}'_{I,J,m_1,m_2} &= \sum_{l,m}\mathbf{C}^{l,m}_{l_1=1,m_1,l_2=1,m_2}\mathbf{\tilde{h}}_{I,J,l,m}
% \end{align}
% where $\mathbf{C}^{l,m}_{l_1,m_1,l_2,m_2}$ are the Clebsch-Gordon coefficients ensuring that the relation in \eqref{eq:hessian_symmetry} is enforced \citep{unke2021se}. $\mathbf{H}'_{I,J,m_1,m_2}$ is a $3\times3$ block, since $m_1$ and $m_2$ run from $0$ to $2$. $\mathbf{W}_c$ is a linear layer contracting over the channel dimension to get a single irrep of degree $\texttt{1x0e + 1x1e + 1x2e}$. Since two irreps of degree $l=1$ can only couple to output irreps of degree up to $l=2$, $\mathbf{h}_{I, J,l,m,c}$ only needs to contain irreps up to $l=2$ to build the $3\times3$ Hessian sub-blocks.
Finally, we symmetrize the intermediate Hessian to the final prediction with $\mathbf{H}=\mathbf{H}' + \mathbf{H}'^\top$.

\subsection{Loss function design}
To learn the Hessians, one can use standard loss functions like mean absolute error (MAE) or mean squared error (MSE) between predicted and actual Hessian elements:
\begin{align}
    \mathcal{L}_\text{MAE / MSE} = \sum_{i,j} |\mathbf{H}_{i,j} - \mathbf{H}^\text{pred}_{i,j}|_\text{MAE / MSE}
\end{align}
However, we are often mainly interested in the smallest eigenvalues and the corresponding eigenvectors \citep{RSRFO1998a, cerjanFindingTransitionStates1981}. 
%For this reason, we design a loss function to emphasize this subspace of the eigen-decomposition.
For this reason, we use a loss function that emphasizes the relevant subspace of the Hessian. This loss function is similar to \cite{li2025enhancing}, which uses it to emphasize the subspace up to the Lowest Unoccupied Molecular Orbital (LUMO) in Fock matrix prediction.
Let $\mathbf{V} = [\mathbf{v}_1,\mathbf{v}_2,...,\mathbf{v}_{3N}]$ be the matrix with columns made from eigenvectors of the ground truth Hessian $\mathbf{H}$, and $\mathbf{\Lambda}$ the corresponding matrix with eigenvalues on the diagonal and zeros everywhere else. Further denote $\mathbf{V}_{[:,:k]}$ and $\mathbf{V}_{[:,k:]}$ the matrices sliced to contain all columns up to/starting from the $k^{th}$ one. We can then define a subspace loss by
\begin{align}
    \mathcal{L}_\text{sub} = \sum_{i,j} \left|\mathbf{V}_{[:,:k]}^\top \mathbf{H}^\text{pred}\mathbf{V}_{[:,:k]} - \mathbf{\Lambda}_{[:,:k]}\right|_{i,j} 
    % + \gamma \sum_{i,j} \left|\mathbf{V}_{[:,k:]}^\top \mathbf{H}^\text{pred}\mathbf{V}_{[:,k:]} - \mathbf{\Lambda}_{[:,k:]}\right|_{i,j}
    \label{eq:eigenloss}
\end{align}
If $\mathbf{H}^\text{pred} = \mathbf{H}$ we have $\mathbf{V}^\top \mathbf{H}^\text{pred}\mathbf{V} = \mathbf{\Lambda}$, and the loss has therefore the correct minimum. Using $\mathcal{L}=\mathcal{L}_\text{MAE/MSE}+\alpha\mathcal{L}_\text{sub}$ lets us emphasize the subspace corresponding to the $k$ lowest eigenvectors and eigenvalues. Since the Hessian has five or six redundant translational and rotational degrees of freedom with eigenvalues close to zero, we need to set $k>6$. In practice, we use $k=8$.
% \\
As we will show in \ref{sec:lossablation}, predicting Hessians with HIP using any loss function significantly outperforms AD Hessians. MAE combined with the subspace loss in \eqref{eq:eigenloss} further improves results in transition state search and extrema classification with frequency analysis by a small margin, which is why we use them in the experiment of the main text. 
For details and ablations of the choice of loss function see \ref{sec:lossablation}.

\section{Experiments\label{sec:experiments}}
We evaluate our proposed approach for direct prediction of molecular Hessians across multiple tasks that test both accuracy and practical utility. Specifically, we measure prediction accuracy and quantify computational speed-ups relative to AD and finite-difference Hessian calculation. We further assess downstream performance in geometry optimization, zero-point energy, and TS searches. For TS searches, we additionally perform frequency analyses to evaluate the reliability of the predicted Hessians in distinguishing true transition states. 
%Together, these experiments provide a comprehensive picture of both the efficiency and usability in common chemical modeling tasks. 

\subsection{Dataset and training}
We evaluate HIP on the Hessian dataset for Optimizing Reactive MLIP (HORM) \citep{cui2025horm}, comprising 1,725,362 training geometries from Transition1x (T1x) \citep{schreiner2022transition1x}, 50,844 T1x validation geometries, and 60,000 HORM-RGD1 geometries used for size-generalization tests.
All HORM energy, force, and Hessian labels were computed with GPU4PySCF v1.3.0 \citep{wu2025gpu4pyscf} using the $\omega$B97X functional \citep{chai2008wb97x} and the 6-31G* basis set \citep{ditchfield1971basis,hariharan1973polarization}, equivalently denoted 6-31G(d).
Molecules are closed-shell neutrals with charge 0 and spin multiplicity 1.
The self-consistent field was converged to an energy change below $10^{-10}$~Hartree and an orbital-gradient norm below $10^{-5}$, the GPU4PySCF v1.3.0 defaults.
Analytical Hessians were obtained from the PySCF \texttt{Hessian()} implementation.
Additional DFT calculations performed for this work use the same $\omega$B97X/6-31G* level unless noted: PubChem size-extrapolation references with ORCA~6.0 \citep{ORCA} (\ref{sec:pubchemorca}), glycine proton transfer with ORCA~6.1.1 at $\omega$B97X-D3/6-31G(d) (\ref{sec:glycinedetails}), and ReactBench DFT verification with GPU4PySCF v1.3.0 (\ref{sec:reactbench}).\\
To isolate Hessian prediction quality from energy and force predictions, we train only the Hessian readout head while keeping the rest of EquiformerV2, pre-trained on energy, forces, and AD Hessians, frozen. This ensures that HIP-EquiformerV2 and AD-EquiformerV2 have identical energy and force predictions across all experiments. All AD-Hessian baseline models were trained on the same ground-truth DFT Hessians, so improvements are purely due to the advantage of HIP prediction over AD.

% \subsection{Accuracy}
\subsection{Accuracy}
We assess Hessian quality using the element-wise mean absolute error (MAE), the MAE of the eigenvalues $\lambda$ and the smallest eigenvalue $\lambda_1$, and the cosine similarity of the smallest eigenvector $\bm{v}_1$. We focus particularly on the lowest eigenvalue and eigenvector as they play an important role in downstream applications such as transition state search. We also measure the average inference time. Table~\ref{tab:model-performance} shows results on the HORM-T1x validation set.
The $\lambda_1$ and $\bm{v}_1$ metrics are computed after mass-weighting and Eckart projection onto the vibrational subspace.
Models trained only on energy and force labels, such as EquiformerV2 (E-F), have substantially higher Hessian errors, showing that accurate second derivatives require explicit Hessian training. Among the methods using automatic differentiation with Hessian training data, EquiformerV2 has the lowest errors. HIP-EquiformerV2 improves on all baselines across every reported metric. Compared with AD-EquiformerV2, HIP has $2.5\times$ lower Hessian MAE, $3.8\times$ lower eigenvalue MAE and higher eigenvector cosine similarity, while being $16\times$ faster. End-to-end HIP training (marked by * in Table~\ref{tab:model-performance}) reduces the errors further, reaching a Hessian MAE of $0.020$ eV/\AA$^2$ at $31.4$ ms per prediction.

\begin{table}[t]
\begin{tabular}{lccccccc}
\hline
\multirow{2}{*}{Model} & Hessian & Hessian & $H$ $\downarrow$ & $\lambda$ $\downarrow$ & Cos $\bm{v}_1$ $\uparrow$ & $\lambda_1$ $\downarrow$ & Time $\downarrow$ \\
 & Method & Trained & eV/\AA$^2$ & eV/\AA$^2$ & unitless & eV/\AA$^2$ & ms \\
\hline
AlphaNet (E-F) & \multirow{4}{*}{AD} & & 0.502 & 1.190 & 0.903 & 0.245 & 728.6 \\ 
LEFTNet-DF (E-F) & & & 1.650 & 2.247 & 0.505 & 1.362 & 331.4 \\
LEFTNet-CF (E-F) & & & 0.364 & 1.011 & 0.947 & 0.130 & 1047.9 \\
EquiformerV2 (E-F) & & & 2.254 & 4.199 & 0.279 & 1.372 & 564.9 \\
\hline
AlphaNet & \multirow{4}{*}{AD} & \checkmark & 0.390 & 0.790 & 0.899 & 0.244 & 747.0 \\
LEFTNet-DF & & \checkmark & 0.200 & 0.172 & 0.937 & 0.136 & 332.2 \\
LEFTNet-CF & & \checkmark & 0.153 & 0.226 & 0.951 & 0.127 & 1051.6 \\
EquiformerV2 & & \checkmark & 0.077 & 0.070 & 0.916 & 0.104 & 562.3 \\
\hline
HIP-EquiformerV2 & \multirow{2}{*}{HIP} & \checkmark & 0.030 & 0.063 & \textbf{0.982} & \textbf{0.031} & 38.5 \\
HIP-EquiformerV2* & & \checkmark & \textbf{0.020} & \textbf{0.041} & \textbf{0.982} & \textbf{0.031} & \textbf{31.4} \\
\hline
\end{tabular}
\caption{
\textbf{Hessian prediction errors on the HORM-T1x validation set.} Denoted are the mean absolute errors (MAE), cosine similarity (Cos), and average time per forward pass on an RTX 3060 GPU. $\lambda_1$ and $\bm{v}_1$ were obtained after Eckart-projection.
Bold values highlight the best-performing model. Baseline AD models are taken from \cite{cui2025horm}. In HIP-EquiformerV2 we trained only the Hessian readout head while keeping the parameters of the backbone frozen, while HIP-EquiformerV2* is trained end-to-end.
}
\label{tab:model-performance}
\end{table}

\subsection{Glycine proton transfer}
To assess Hessian quality along a chemically meaningful motion, we study the intramolecular proton transfer of glycine from the Transition1x test split \cite{schreiner2022transition1x}. We describe the transferring proton using its N--H and O--H distances, \(q_{\mathrm{NH}}\) and \(q_{\mathrm{OH}}\), and define the proton-transfer coordinate \(\xi=q_{\mathrm{NH}}-q_{\mathrm{OH}}\) (see Sec.~\ref{sec:glycinedetails}).

\begin{figure}
    \centering
    \includegraphics[width=0.99\linewidth]{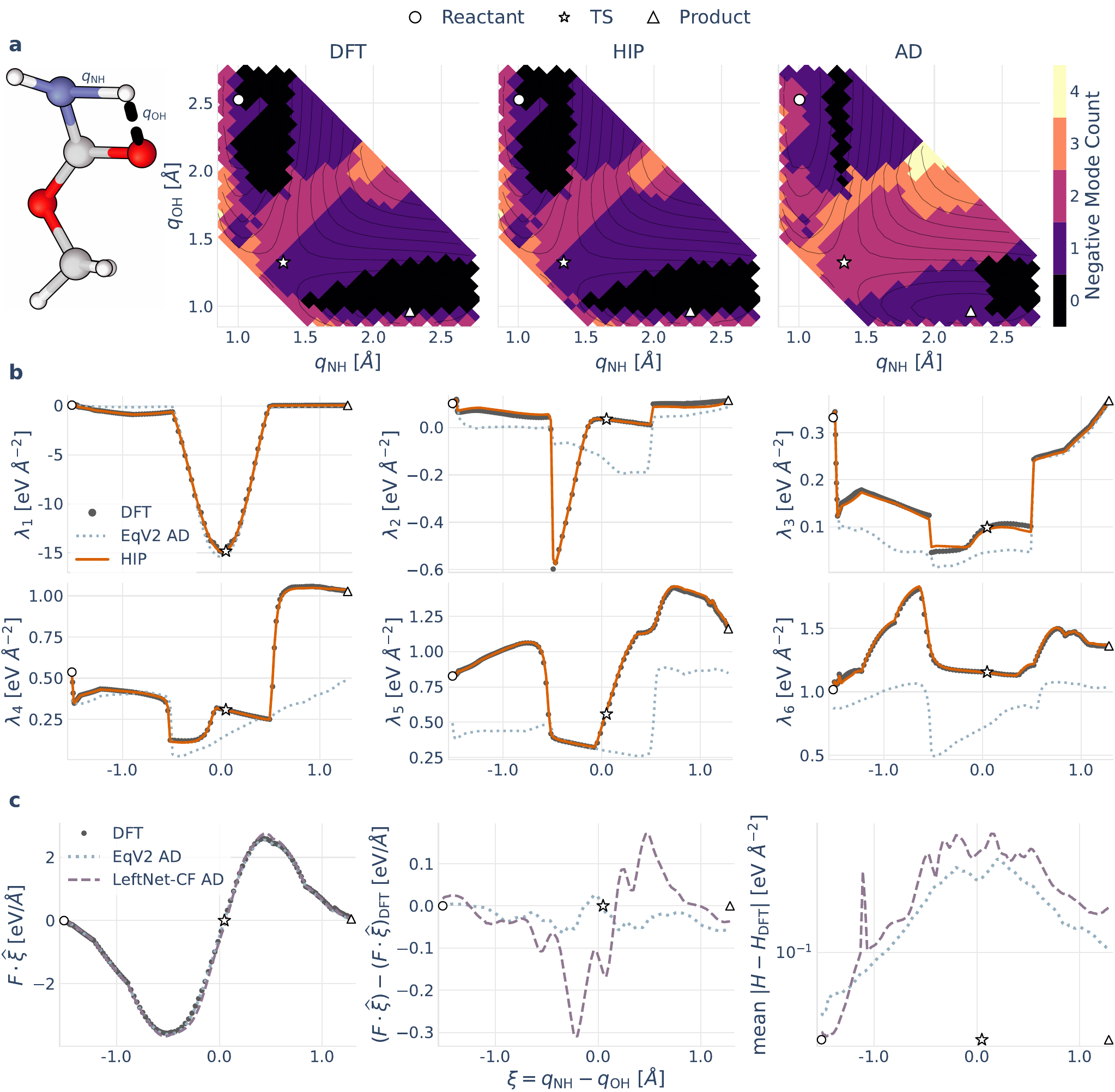}
    \caption{
    \textbf{Glycine proton transfer.}
    \textbf{(a)} Number of negative eigenvalues of the Eckart-projected, mass-weighted Hessian over the two-dimensional surface defined by the N--H and O--H distances, $q_{\mathrm{NH}}$ and $q_{\mathrm{OH}}$. DFT is the reference; HIP is direct Hessian prediction; AD denotes Hessians differentiated from learned forces. Shading is the integer negative-mode count and black lines are DFT energy contours. Circles, stars and triangles mark the reactant, transition state and product. In the rendering, carbon is grey, hydrogen white, nitrogen blue and oxygen red.
    \textbf{(b)} Six lowest Hessian eigenvalues along 145 minimum-energy-path (MEP) geometries versus $\xi=q_{\mathrm{NH}}-q_{\mathrm{OH}}$. DFT is shown as grey points, HIP as solid orange lines and EquiformerV2 (EqV2) AD as dotted blue lines.
    \textbf{(c)} Force projected onto the unit tangent $\widehat{\bm{\xi}}$ of the MEP (left), force residual relative to DFT (middle) and Hessian MAE (right). DFT is shown as grey points, EqV2 AD as dotted blue lines and LeftNet-CF AD as solid purple lines.
    }
    \label{fig:glycine}
\end{figure}

We first study the two-dimensional \((q_{\mathrm{NH}}, q_{\mathrm{OH}})\) surface around the proton-transfer transition state. At each grid point, we place the proton at the target distances and relax all remaining degrees of freedom while keeping both distances constrained. The resulting geometries are then reoptimized at the DFT level. We characterize the surface by counting the negative eigenvalues
% of the Eckart-projected, mass-weighted Hessian from DFT, HIP, and AD Hessian calculations 
(Fig.~\ref{fig:glycine}a). HIP closely reproduces the DFT topology, whereas AD introduces spurious negative modes near both the transition state and minima.

We next track the six lowest vibrational eigenvalues along a fixed, all-atom minimum-energy path as a function of \(\xi\) (Fig.~\ref{fig:glycine}b). HIP closely follows the DFT reference for all six modes. AD Hessians reproduce the lowest mode more closely but systematically underestimate the orthogonal curvatures of modes~2--6. This agrees with Table~\ref{tab:model-performance}, where HIP captures curvature information that AD Hessians miss.

Finally, we project the Cartesian force \(\mathbf{F}\) onto the unit tangent \(\widehat{\bm{\xi}}\) of the minimum-energy path, \(\mathbf{F}\cdot\widehat{\bm{\xi}}\) (Fig.~\ref{fig:glycine}c). We compare EquiformerV2 (EqV2) and LeftNet with conservative forces (LeftNet-CF), differentiating both models to obtain AD Hessians. Although both MLIPs reproduce the overall DFT force trend (left), their projected forces oscillate relative to DFT (middle). Differentiation amplifies these fluctuations into large Hessian errors along the path (right). Together with the spurious negative modes and underestimated orthogonal curvatures, this is consistent with AD Hessian errors arising from noise in the learned MLIP potential-energy surface.
% \[
% (F_{\mathrm{model}}\cdot\widehat{\xi}) - (F_{\mathrm{DFT}}\cdot\widehat{\xi})
% = (F_{\mathrm{model}} - F_{\mathrm{DFT}})\cdot\widehat{\xi}.
% \]

\subsection{Computational efficiency and scaling\label{sec:speedup}}
\begin{figure}
    \centering
    \includegraphics[width=0.99\linewidth]{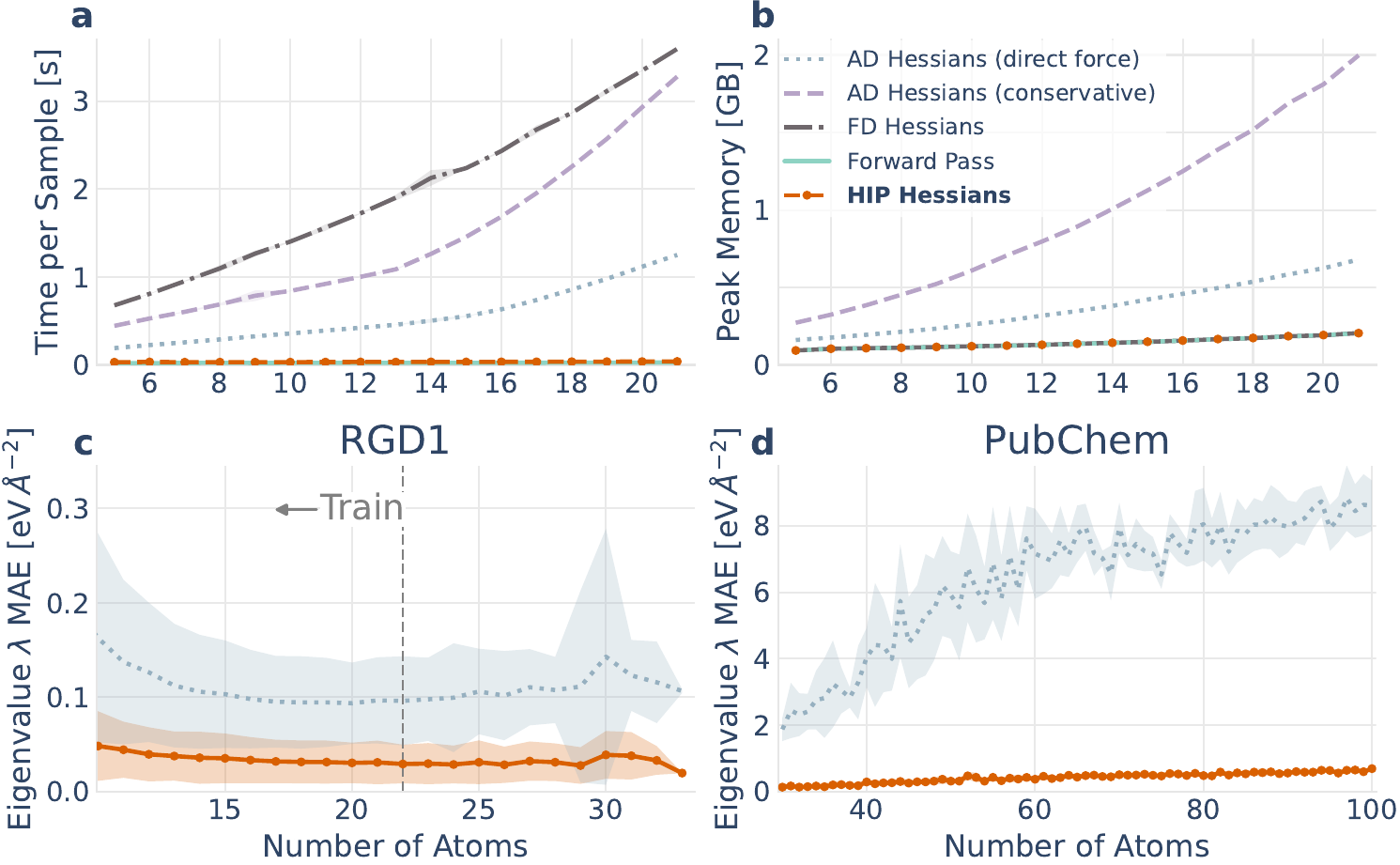}
     \caption{\textbf{Computational cost and molecular-size scaling.}
     HIP is direct Hessian prediction; AD denotes Hessians differentiated from learned forces or energies; finite difference (FD) denotes numerical force differentiation.
     \textbf{(a,b)} Mean single-molecule inference time (a) and peak GPU memory (b) versus atom count on an RTX 3060 GPU; shading is one standard deviation over replicates. The forward pass is teal and dotted, HIP orange, direct-force AD blue, conservative-force AD purple and FD dark grey. The forward-pass and HIP curves overlap.
     \textbf{(c,d)} Mean eigenvalue MAE of the Eckart-projected Hessian versus atom count for HORM-RGD1 (c) and ten PubChem samples per atom count (d) \citep{kim2025pubchem}; shading is one standard deviation over samples. HIP is shown by orange diamonds and AD by blue circles. The vertical grey dashed line in panel (c) marks the largest molecule size in the training set.
     }
     \label{fig:speed_memory}
\end{figure}

% The results in Table \ref{tab:model-performance} showed a drastic improvement in average prediction time compared to all other models using automatic differentiation. 
% In Figure \ref{fig:speed_memory}(a-b), we further compare the prediction time and memory footprint as a function of molecule size. The direct prediction of Hessians is $10$ - $74\times$ faster than using AD for molecules in the HORM dataset. HIP predicted Hessians also benefit from a much more favourable scaling with the molecule size. Figure \ref{fig:speed_memory}(c) compares the batched AD Hessian as implemented in \citep{cui2025horm} with our HIP-EquiformerV2. We observe at minimum a $78\times$ speedup as a function of batch size compared to AD. We explain the significant performance degradation of batched AD Hessians in \ref{sec:batched_ad_explanation}.
A critical bottleneck in materials discovery is the scaling of compute with system size.
Fig. \ref{fig:speed_memory}a-b compares the inference time and memory footprint of HIP versus AD approaches.
HIP exhibits the same scaling as a regular forward pass, with only a constant 25\% additional time, roughly the cost of one extra layer on top of the four hidden layers.
AD approaches, on the other hand, suffer from having to backpropagate through the graph for each row of the Hessian, which increases the cost to $3N$ of a forward pass.
Quantitatively, HIP yields a $10\times$ to $70\times$ speedup on small molecules (5-30 atoms) and reduces peak memory usage by a factor of 2-3 compared to AD.
This efficiency increases the size of systems for which Hessian-based workflows are computationally practical.
AD calculations are also much harder to parallelize across a batch dimension; see \ref{sec:batched_ad_explanation} for the explanation. 
By comparison, since HIP reuses the message passing machinery, HIP parallelizes just as easily as the forward energy pass. We compared the batched AD Hessian as implemented by Cui et al.~\citep{cui2025horm} with our HIP-EquiformerV2. 
At the largest evaluated HIP batch size, the per-sample inference time is at least $81\times$ lower than that of their batch-one AD Hessian implementation; see \ref{sec:batched_ad_explanation} for more details.

\subsubsection{Generalization to larger systems}
To assess extrapolation beyond the training distribution, we evaluate HIP on HORM-RGD1, using the published GPU4PySCF $\omega$B97X/6-31G* labels, and on 710 PubChem samples containing 30--100 atoms (10 samples per atom count) \citep{kim2025pubchem}, with reference Hessians computed with ORCA~6.0 at the same level (\ref{sec:pubchemorca}). Fig.~\ref{fig:speed_memory}c-d shows the error of the Hessian eigenvalues as a function of the number of atoms, after mass-weighting and Eckart projection (\ref{sec:pubchemorca}). We use the eigenvalues as a metric since their scale is independent of system size, while the MAE of Hessian elements lead to misleadingly low errors as a large share of Hessian elements go to zero for larger systems. 
HIP maintains strong performance across system sizes, with errors remaining relatively flat even for molecules larger than those in the training set. 
% The favorable $O(N)$ scaling of the sparse Hessian prediction ensures that both accuracy and computational efficiency extend to larger systems. 
In contrast, the AD-based predictions show degraded performance along with the prohibitive computational costs for large molecules.

\subsection{Validating downstream utility}

\begin{figure}
    \centering
    \includegraphics[width=0.99\linewidth]{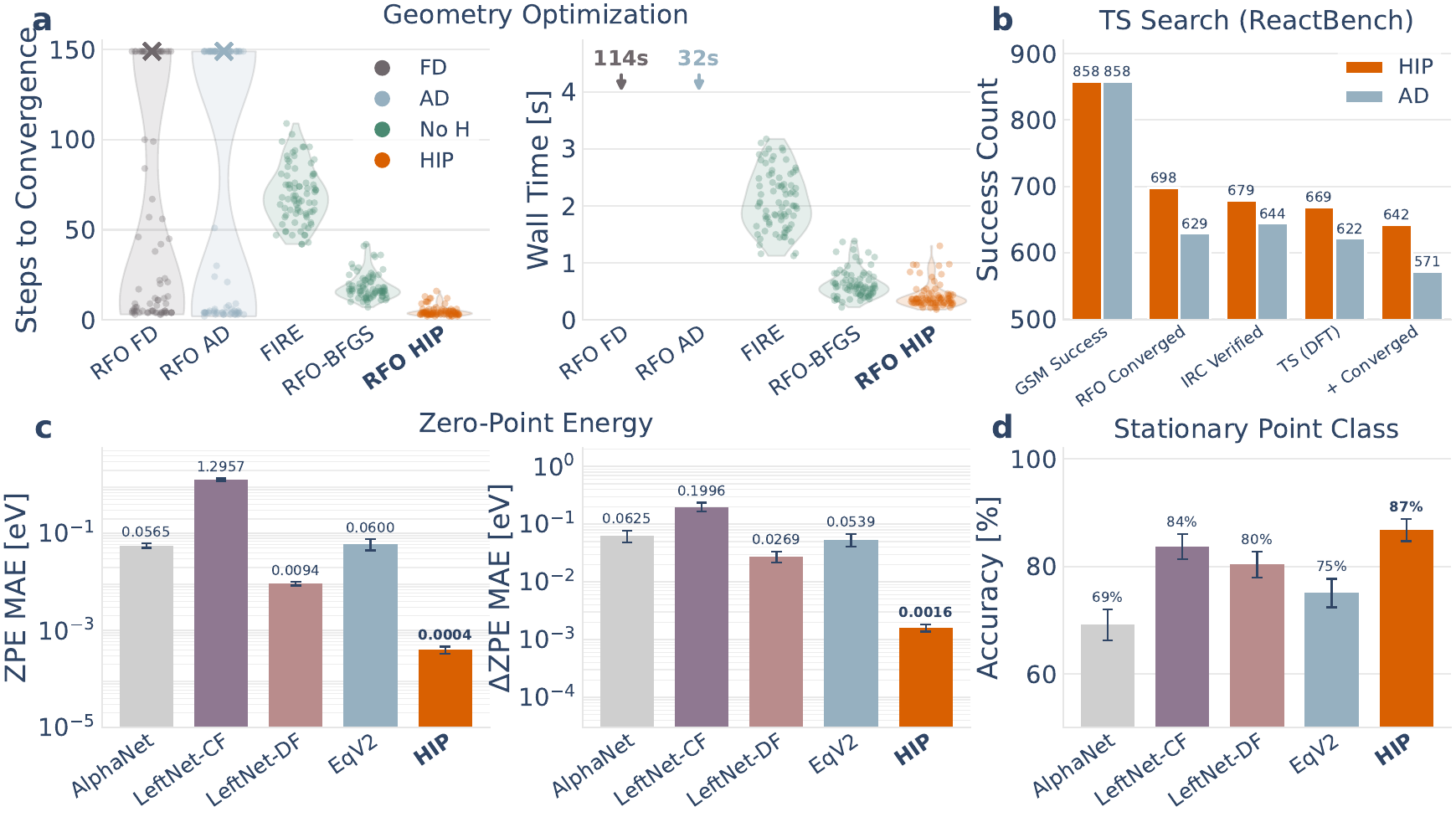}
    \caption{
    \textbf{Downstream applications.}
    HIP is direct Hessian prediction; AD denotes Hessians differentiated from learned forces; DFT is the reference.
    \textbf{(a)} Optimization steps and wall-clock time for 80 geometries. Violin shading is the sample distribution, points are individual geometries and crosses mark runs that did not converge within 150 steps. RFO wall-clock times with AD or finite-difference Hessians are omitted because they are one to two orders of magnitude longer than RFO with HIP.
    \textbf{(b)} Successful ReactBench samples after GSM, RFO, IRC, DFT transition-state verification and an additional tight force-convergence criterion, from 960 initial reactions \citep{zhao2025}. DFT-verified full-set counts are extrapolated from fixed subsets of 100 TS proposals per Hessian method.
    \textbf{(c)} Mean absolute reactant ZPE error and reaction $\Delta$ZPE error relative to DFT over 80 reactant--product pairs; the vertical axis is logarithmic.
    \textbf{(d)} Fraction of 1{,}000 stationary points for which the exact number of negative Hessian eigenvalues is predicted. Error bars in (c) and (d) are 95\% bootstrap confidence intervals.
    }
    \label{fig:relaxations_tssearch_zpe}
\end{figure}

\subsubsection{Geometry optimization \label{sec:relax}}
We evaluate HIP Hessians in geometry optimization workflows, comparing exact second-order methods (rational function optimization, RFO) using Hessians against first-order methods and quasi-second-order RFO using BFGS.
Fig.~\ref{fig:relaxations_tssearch_zpe}a shows convergence statistics across the validation set. RFO with HIP Hessians converges in the fewest steps on average, whereas finite-difference and AD Hessians often fail to converge within a 150-step budget. Na\"ive steepest descent rarely converges, underscoring the importance of curvature information.
Wall-clock timings in Fig.~\ref{fig:relaxations_tssearch_zpe}a show the same trend. RFO with HIP Hessians is fastest overall because it requires fewer optimization steps and less expensive Hessian evaluations, whereas AD and finite-difference approaches are orders of magnitude slower.
% The current implementation is bottlenecked by memory transfer from and to the CPU. We would expect an even more significant speedup if the optimization was performed on GPU and parallelized across samples.

\subsubsection{Transition state search}
Transition state search is a central challenge in computational chemistry, requiring accurate Hessian eigenvalues and eigenvectors. We evaluate HIP using the ReactBench benchmark \cite{zhao2025} with additional DFT verification (\ref{sec:reactbench}). The workflow comprises four stages: the growing string method (GSM) first approaches the TS using energies and forces; RFO optimization with Hessians then refines the TS; the intrinsic reaction coordinate (IRC) then verifies connectivity; and a final DFT Hessian confirms the saddle point.
Fig.~\ref{fig:relaxations_tssearch_zpe}b shows the success rates at each stage.\\
% Total Success: $89.38\%$ ($858/960$)
% RFO Convergence: $72.71\%$ (HIP) vs $65.52\%$ (AD)
% IRC Verification: $70.73\%$ (HIP) vs $67.08\%$ (AD)
% DFT-Verified Transition States: $69.69\%$ (HIP) vs $64.79\%$ (AD)
GSM performs equally well for both models because it relies only on energies and forces, which are identical. The differences emerge in Hessian-dependent stages. HIP achieves higher RFO convergence, better IRC verification, and most importantly, more DFT-verified transition states. When requiring both DFT verification and tight convergence, HIP improves 12\% over AD.
These results demonstrate that HIP's superior Hessian quality translates to more reliable TS identification, which is crucial for estimating reaction barriers and rates.

\subsubsection{Zero-point energy}
% Absolute ZPE (Error Reduction by HIP):
% vs. LeftNet-DF: HIP reduces error by $95.74\%$.
% vs. AD-EquiformerV2: HIP reduces error by $99.33\%$.
% Relative $\Delta$ZPE (Error Reduction by HIP):
% vs. AD-EquiformerV2: HIP reduces error by $97.03\%$.
Zero-point energy is a critical thermochemical property derived from vibrational frequencies, which depend directly on Hessian eigenvalues. We evaluate both absolute ZPE at the reactant geometry and relative $\Delta$ZPE between reactants and products (see~\ref{sec:zpe_setup}).
Fig.~\ref{fig:relaxations_tssearch_zpe}c reveals dramatic improvements with HIP. For absolute ZPE, HIP achieves an order of magnitude better accuracy than LeftNet-DF and two orders of magnitude better than AD-EquiformerV2. For relative $\Delta$ZPE, HIP similarly reduces the error by $97.03\%$ compared to AD-EquiformerV2.
These results demonstrate that HIP's improved eigenvalue predictions translate directly to better thermochemical properties, where small errors in vibrational frequencies can lead to large errors in free energy.

\subsubsection{Frequency analysis for extrema classification}
Characterizing stationary points on potential energy surfaces requires determining the number of negative Hessian eigenvalues: zero for minima, one for first-order transition states, and two or more for higher-index saddle points \cite{maronsson2012method}. We test this capability on 1000 geometries from the HORM-T1x validation, comprising 44\% first-order TSs, 37\% minima, and 19\% higher-order saddle points, as determined from DFT Hessians.
Fig.~\ref{fig:relaxations_tssearch_zpe}d shows classification accuracy based on counting negative eigenvalues. HIP achieves 92\% accuracy, 10 percentage points better than the next-best AD model and 17 percentage points better than AD-EquiformerV2. This substantial improvement stems from HIP's more accurate eigenvalue predictions, particularly for the small eigenvalues that determine the character of the stationary point.
Accurate extrema classification is essential for automated reaction pathway discovery and for validating that optimizations have reached the intended type of stationary point.

\subsection{Synergistic improvements from Hessian training}
\begin{figure}
    \centering
    \includegraphics[width=0.99\linewidth]{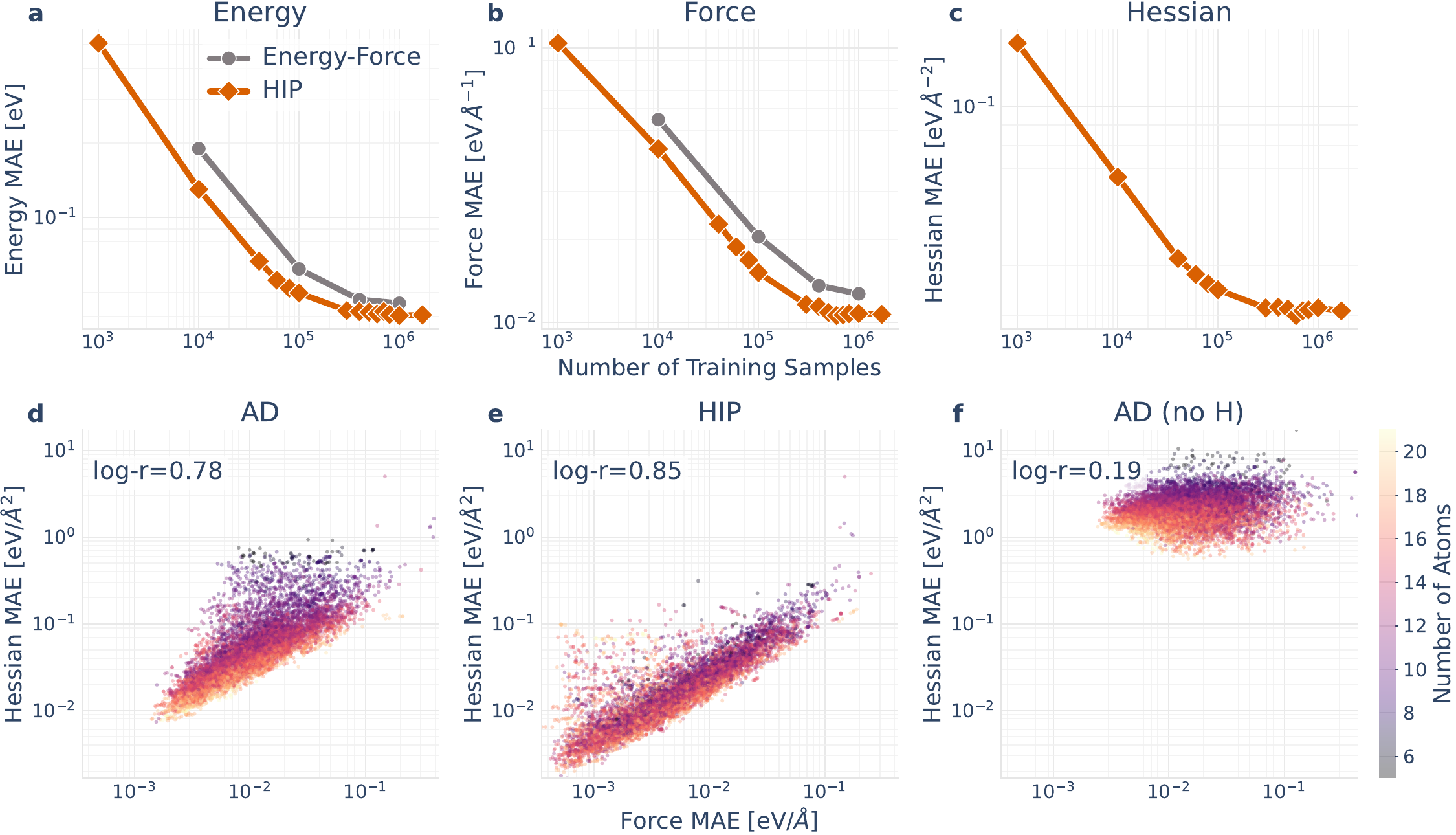}
     \caption{
     \textbf{Dataset scaling and force--Hessian error correlation.}
     MAE is evaluated on the HORM-T1x validation set.
     \textbf{(a--c)} Energy (a), force (b) and Hessian (c) MAE versus the number of randomly selected training samples. EquiformerV2 trained on energies and forces is shown in dark grey; HIP trained end-to-end on energies, forces and Hessians is shown in orange.
     \textbf{(d--f)} Per-geometry force and Hessian MAE for 10{,}000 validation geometries using EquiformerV2 with Hessian supervision and AD Hessians (d), HIP trained end-to-end (e) and EquiformerV2 trained only on energies and forces with AD Hessians (f). Each point is one geometry. Colour indicates atom count. The displayed log-$r$ is the Pearson correlation between the base-10 logarithms of the force and Hessian errors.
     }
     \label{fig:datascaling}
\end{figure}

Thus far, we trained only the Hessian readout while freezing the backbone, isolating Hessian quality from energy and force predictions. We now investigate joint end-to-end training of the full model, including Hessian supervision.
We train both the standard EquiformerV2 (energy and forces only) and the HIP-EquiformerV2 (energy, forces, and Hessians) for equal wall-clock time (3 days) across various dataset sizes. Fig.~\ref{fig:datascaling}a-c reveals that Hessian supervision improves not only Hessian predictions but also energy and force accuracy across all data regimes.

On 10{,}000 validation structures (Fig.~\ref{fig:datascaling}d-f), force and Hessian errors are positively correlated for Hessian-trained models, but nearly independent for an energy-force-only baseline, which can achieve good force accuracy while Hessian MAE remains an order of magnitude higher. HIP occupies a tighter, lower-error region than AD, indicating a more self-consistent potential energy surface.
%
% Fig.~\ref{fig:datascaling}(d-f) shows that Hessian supervision couples force and curvature accuracy across geometries, whereas energy-force training alone leaves large Hessian errors largely uncorrelated with force error.
%
% Fig.~\ref{fig:datascaling}(d-f) plots per-structure force and Hessian MAE on 10{,}000 HORM-T1x validation geometries. Models trained with Hessian labels exhibit a clear positive correlation between the two errors; an energy-force-only model does not, despite comparably low force errors on many structures. HIP lies below and tighter than AD, combining lower errors with stronger force-Hessian consistency.

This synergistic effect suggests that Hessian information provides a richer training signal, helping the model learn a better potential energy surface representation.

\paragraph{Non-Conservative Hessians}
The practical implication of HIP is that second-order information can be used more routinely in computational workflows where its cost has previously been prohibitive.
More broadly, expensive tensorial quantities with known symmetries may be represented by dedicated equivariant readouts, removing the need for costly differentiation of a learned scalar potential.

Direct prediction, however, does not guarantee that HIP Hessians are globally integrable or exactly consistent with the predicted energies and forces. Unlike non-conservative force errors, which can cause systematic energy drift during long molecular dynamics trajectories, Hessian errors in the applications considered here enter as local curvature estimates. This is consistent with the widespread use of approximate Broyden--Fletcher--Goldfarb--Shanno (BFGS) Hessians, which indicates that exact consistency is not necessary for all workflows (Appendix, \ref{sec:conservative}). Applications requiring exact energy--force--Hessian consistency may therefore need additional architectural or loss-based constraints.

\paragraph{Limitations}
Other limitations are shared with existing MLIPs. Accuracy depends on the chemical and configurational coverage of the training data, and the finite interaction cutoff may omit long-range curvature contributions in ionic, polar or extended systems. Moreover, generating reference Hessians makes broad training coverage more expensive than for energy- and force-only models.

One possible way to reduce this data cost is to bootstrap HIP with approximate Hessian labels obtained by automatic differentiation of an energy-and-force model, followed by fine-tuning on a smaller set of reference DFT Hessians \cite{gardner2024}. The utility of such pseudo-label pre-training will depend on whether fine-tuning can correct systematic errors and noise in the initial AD Hessians, including the curvature errors observed here. Quantifying the trade-off between pseudo-label quality and the amount of reference Hessian data is therefore an important direction for future work.

Future experiments should validate HIP across broader chemical spaces and higher-fidelity electronic-structure methods. For periodic systems, the local construction can be adapted using a periodic neighbour list. The real-space supercell Hessian could then be Fourier transformed into the momentum-dependent dynamical matrix. The same symmetry-based, direct-prediction approach could also be extended to higher-rank response tensors. For example, the elastic tensor would require an additional global rank-4 readout containing $l=4$ features, while predicting the cubic force constant tensor would require an atom-triplet rank-3 readout containing $l=1$ and $l=3$ features.

\section{Conclusion}
In this work, we have presented Hessian Interatomic Potentials (HIP), a method for directly predicting molecular Hessians using SE(3)-equivariant neural networks. Our approach eliminates the need for traditional computationally expensive methods, such as finite differences or automatic differentiation, offering a significant improvement in computational time, memory usage, and accuracy.
Through extensive validation across a range of molecular tasks, 
% such as transition-state search, geometry optimization, zero-point energy corrections, and vibrational frequency analysis, 
we have shown that our predicted Hessians are highly effective for practical applications in computational chemistry. 

Due to the versatility of the framework, any base model using spherical harmonics and message passing can be made HIP.

% Initial idea by L.T.. Code, training, experiments, and figures by A.B.. Paper writing by L.T., A.B., N.R..

\subsubsection*{Author contributions}
Code, training, and experiments by A.Bu.; idea, theory, and Clebsch--Gordan transform code by L.T.; paper writing by L.T., A.Bu.\ and N.R.; visualizations by A.Bu.\ and L.T.; supervision and paper editing by N.R., A.Bh., T.V., V.B.\ and A.A.-G.; manuscript review by A.Bu., L.T., N.R., N.V., V.B., T.V., A.Bh.\ and A.A.-G.

\subsubsection*{Data and code availability}
The HORM training and validation dataset is available from Zenodo at \url{https://doi.org/10.5281/zenodo.17217897} \citep{horm_zenodo,cui2025horm}.
HORM geometries originate from Transition1x \citep{schreiner2022transition1x}, available from figshare at \url{https://doi.org/10.6084/m9.figshare.19614657.v4} \citep{transition1x_figshare}.
DFT calculations generated in this work, including atomic coordinates, energies, forces and Hessians for the PubChem CHNO size-extrapolation set, the glycine proton-transfer surface and the glycine minimum-energy path, together with trained HIP model weights, are available from Zenodo at \url{https://doi.org/10.5281/zenodo.22003643} \citep{hip_zenodo_data} and from Hugging Face at \url{https://huggingface.co/andreasburger/hip}.
The HIP source code is available at \url{https://github.com/BurgerAndreas/hip} under the Apache License, Version 2.0.
A frozen snapshot is archived at Zenodo at \url{https://doi.org/10.5281/zenodo.22003592} \citep{hip_zenodo_code}.
All requests should be addressed to Andreas Burger via \url{me@andreas-burger.com}.

\subsubsection*{Acknowledgements}
We are thankful for funding by the Pioneer Center for Accelerating P2X Materials Discovery (CAPeX), DNRF grant number P3.
% https://www.science.org/doi/10.1126/science.adk9227#acknowledgments
A.A.-G. thanks Anders G. Frøseth for his generous support.
A.B. and L.T. acknowledge the AIST support to the Matter lab for the project titled "SIP project - Quantum Computing".
A.B. is thankful for support through the Google NSERC IRC award.
L.T. thanks the NSERC for the support through the Discovery Grant.
This research was undertaken thanks in part to funding provided to the University of Toronto’s Acceleration Consortium from the Canada First Research Excellence Fund CFREF-2022-00042.
% This research was enabled, in part, by support provided by the SciNet HPC Consortium (scinethpc.ca) and the Digital Research Alliance of Canada. 
% Computations were performed on the Acceleration Consortium AMD Tacozoid cluster.
Resources used in preparing this research were provided, in part, by the Province of Ontario, the Government of Canada through CIFAR, and companies sponsoring the Vector Institute.

\newpage
\bibliography{Hessian}
\bibliographystyle{neurips2025}

%%%%%%%%%%%%%%%%%%%%%%%%%%%%%%%%%%%%%%%%%%%%%%%%%%%%%%%%
\newpage
\appendix
\section{Appendix}
\FloatBarrier

\subsection{Traditional methods for calculating Hessians}
\subsubsection{Analytical Hessians via CPKS}
\label{sec:cpks}
The gold standard for computing nuclear Hessians is to derive them analytically. In Kohn-Sham DFT, this is done with the Coupled Perturbation Kohn-Sham (CPKS) method. This is much more complex and computationally demanding than energy or force calculations. The difficulty arises from the need to calculate the response term of the electron density, which requires understanding how the self-consistent field (SCF) solution to the Kohn-Sham equations changes with changes in nuclear coordinates. To see why this is necessary, write the DFT energy in terms of the density matrix $\mathbf{P}(\mathbf{R}) = (\mathbf{C})^T(\mathbf{R}) \mathbf{S}(\mathbf{R}) \mathbf{C}(\mathbf{R})$ with molecular orbital coefficients $\mathbf{C}$ and overlap matrix $\mathbf{S}$: 
\begin{align}
    E[\mathbf{P}] = \text{Tr}[h\mathbf{P}] + \frac{1}{2} \text{Tr}[\mathbf{J}[\mathbf{P}]] + \rho(\mathbf{P})
\end{align}
To get the force, so the nuclear gradient, we first define the Lagrangian to enforce normalization of the wavefunction:
\begin{align}
\mathcal{L}(C, \epsilon, \mathbf{R}) = E[\mathbf{R}] - \text{Tr}[\epsilon (\mathbf{C}^T\mathbf{S}\mathbf{C} - \mathbf{I})] \label{eq:lagrangian}
\end{align}
We then get the nuclear gradient by differentiating $\mathcal{L}$. The derivative w.r.t. nuclear component $x$ is:
\begin{align}
    \frac{d\mathcal{L}}{dx} &= \frac{\partial\mathcal{L}}{\partial x} + \underbrace{\frac{\partial \mathcal{L}}{\partial \mathbf{C}}}_{=0} \frac{\partial \mathbf{C}}{\partial x} - \text{Tr}\left[\epsilon \mathbf{C}^T \frac{\partial \mathbf{S}}{\partial x} \mathbf{C}\right] + 2 \text{Tr}\left[(\mathbf{S}\mathbf{C}\epsilon)^T\frac{d\mathbf{C}}{dx}\right] \\
    &= \frac{\partial E}{\partial x} - 2 \text{Tr}\left[(\mathbf{S}\mathbf{C}\epsilon)^T\frac{dC}{dx}\right]  -\text{Tr}\left[\epsilon \mathbf{C}^T \frac{\partial \mathbf{S}}{\partial x} \mathbf{C}\right] + 2 \text{Tr}\left[(\mathbf{S}\mathbf{C}\epsilon)^T\frac{d\mathbf{C}}{dx}\right] \\
    &= \frac{\partial E}{\partial x}  -\underbrace{\text{Tr}\left[\epsilon \mathbf{C}^T \frac{\partial \mathbf{S}}{\partial x} \mathbf{C}\right]}_\text{"Pulay terms"} \label{eq:analytic_grad}
\end{align}
Due to the stationarity $\frac{\partial \mathcal{L}}{\partial C} = 0$ we did not need to calculate the density response $\frac{\partial C}{\partial x}$, which is why calculating forces is cheap, on the same order as the cost of energies. However, once we want Hessians, we need to differentiate \eqref{eq:lagrangian} twice:
\begin{align}
    \frac{d^2 \mathcal{L}}{dx dy} &= \frac{\partial^2 \mathcal{L}}{\partial x \partial y} + \frac{\partial^2 \mathcal{L}}{\partial \mathbf{C} \partial x} \frac{d \mathbf{C}}{d y} + \frac{\partial^2 \mathcal{L}}{\partial \epsilon \partial x}\frac{d \epsilon}{dy} \\
    &=  \frac{\partial^2 \mathcal L}{\partial y\,\partial x}
-\begin{bmatrix}
\displaystyle \frac{\partial^2 \mathcal L}{\partial \mathbf{C}\,\partial x} &
\displaystyle \frac{\partial^2 \mathcal L}{\partial \varepsilon\,\partial x}
\end{bmatrix}
\begin{bmatrix}
\displaystyle \frac{\partial^2 \mathcal L}{\partial \mathbf{C}^2} & \displaystyle \frac{\partial^2 \mathcal L}{\partial \mathbf{C}\,\partial \varepsilon} 
\displaystyle \frac{\partial^2 \mathcal L}{\partial \varepsilon\,\partial \mathbf{C}} & \displaystyle \frac{\partial^2 \mathcal L}{\partial \varepsilon^2}
\end{bmatrix}^{-1}
\begin{bmatrix}
\displaystyle \frac{\partial^2 \mathcal L}{\partial \mathbf{C}\,\partial y} 
\displaystyle \frac{\partial^2 \mathcal L}{\partial \varepsilon\,\partial y}
\end{bmatrix}.
\end{align}
Solving the linear system of equations scales formally as $O(N^5)$ in computation and $O(N^4)$ in memory. Similar to energy calculations, density fitting, sparsification (integral screening), and other optimizations can reduce the cost for CPKS. Still, the scaling remains worse in both computation and memory compared to energy calculations. 

\subsubsection{Hessians via finite differences}
\label{sec:finite_difference}
A major downside of the CPKS approach to Hessians is its highly complex implementation and need for exchange-functional-specific derivations. Therefore, many new methods do not support Hessians, and most codes resort to finite-difference calculations in such cases. Using central differences of the forces along the $3N$ Cartesian directions, the Hessian elements are constructed from
\begin{align}
    \mathbf{H}_{ij} \approx \frac{\mathbf{F}_i(\mathbf{R} + h \mathbf{e}_j) - \mathbf{F}_i(\mathbf{R} - h \mathbf{e}_j)}{2h}
\end{align}
This requires $2\times 3N$ displaced \emph{gradient} calculations. Each gradient calculation scales similarly to the energy calculation with $O(N^4)$, leading to a $O(N^5)$, the same as naive CPKS, although typically with more numerical noise and a much larger prefactor because we have to repeatedly converge an SCF, as opposed to solving a large linear system in analytical Hessians.
\subsection{Hessians in Molecular Optimization}

% Since computing the exact Hessian is prohibitively expensive for all but the smallest systems, quasi-Newton methods construct an approximate Hessian that is updated using gradient differences.
% Several widely used algorithms embody different choices of these ingredients. The BFGS method provides a robust quasi-Newton update scheme and is routinely applied in both Cartesian and internal coordinates.

\paragraph{BFGS}
\label{sec:bfgs}
To apply RFO in practice, we need a Hessian at each step. As computing full Hessians is usually too expensive, a common practice is to maintain an approximate Hessian using the BFGS quasi-Newton scheme. BFGS updates approximate Hessians $\mathbf{B}_k$ of the true Hessian $\mathbf{H}(\mathbf{R}_k)$ using the update equation
\begin{align}
    \mathbf{B}_{k+1} = \mathbf{B}_k 
    - \frac{\mathbf{B}_k \mathbf{s}_k \mathbf{s}_k^\top \mathbf{B}_k}{\mathbf{s}_k^\top \mathbf{B}_k \mathbf{s}_k}
    + \frac{\mathbf{y}_k \mathbf{y}_k^\top}{\mathbf{y}_k^\top \mathbf{s}_k},
    \qquad \mathbf{y}_k^\top \mathbf{s}_k > 0,
\end{align}
where the curvature condition $\mathbf{y}_k^\top \mathbf{s}_k>0$ (typically ensured by a Wolfe line search) guarantees that $\mathbf{B}_{k+1}$ remains positive-definite, which is desirable for minimization. In the RFO framework, $\mathbf{B}_k$ simply replaces the exact Hessian $\mathbf{H}$ in the augmented eigenvalue problem.

\subsubsection{Transition State Search \label{sec:tssearchliterature}}
%Why are they useful, details
%Transition state search methods, growing string method, gentlest ascend dynamics, notch elastic band
Transition states are first-order saddle points on the PES. They represent the maximum barrier along the minimum energy path (MEP) between two minima. Identifying transition states is crucial for understanding reaction mechanisms and mapping reaction networks. A first-order saddle point is characterized by having exactly one negative eigenvalue of the Hessian. Over the years, various computational methods have been developed to describe MEPs and locate transition states. \\
Computational methods for transition state search can broadly be categorized as \textit{single-ended} and \textit{double-ended}. In double-ended methods, we know the product and reactant states (the two minima on the energy surface) and seek an interpolation along the MEP that passes through the TS. In contrast, in single-ended methods, we only know a starting point and climb the energy surface to find a nearby transition state. Often, double-ended methods are used to give good initial guesses, which are then refined by single-ended methods. In this study, we use the Growing String method to find the initial states, which we then refine using RS-P-RFO (see above).\\
Different double-ended methods are distinguished by how they grow the interpolations between product and reactant states. The most common approaches are the nudge elastic band method \citep{henkelman1999, henkelman2000} and the growing string method (GSM) \citep{peters2004,zimmerman2015}. 
% Here, we only review the growing string method as we use it in our experiments for the initial guesses of the transition states.

The nudged elastic band method aims to find the minimum energy path between given initial and final configurations. A set of images connecting these states is linked by classical spring forces, forming an elastic band. Each image experiences a total force comprising the spring force along the local tangent and the true force perpendicular to it. The images are simultaneously optimized to trace the MEP. To converge to the transition state, a climbing image is introduced, for which the force along the band is inverted, leaving only the perpendicular component of the true force. This drives the climbing image up the potential energy surface along the band while descending perpendicularly, reaching the transition state along the MEP.

The growing string method (GSM) locates transition states by incrementally constructing a discrete path of structures on the potential energy surface. Starting from the reactant and product endpoints, two path fragments are iteratively grown together. At each iteration, a new node on the string is added along the local tangent direction and relaxed orthogonally to the path according to the force
\begin{equation}
\mathbf{F}_{\bot} = \mathbf{F} - (\hat{\mathbf{t}}^{\top} \mathbf{F}) \hat{\mathbf{t}},
\end{equation}
where $\mathbf{F} = -\nabla E$ is the force and $\hat{\mathbf{t}}$ is the local tangent unit vector. Convergence is typically accelerated by approximate Hessians, constructed in delocalized internal coordinates and updated by quasi-Newton schemes, which enable eigenvector-guided optimization while avoiding full second-derivative evaluations \citep{zimmerman2015}. Once the two fragments merge, the full string is reparameterized to maintain uniform node distance, and the highest-energy node serves as a transition state estimate, which can be further refined using Hessian-based eigenvectors \citep{peters2004}.

\subsection{Hessian properties}
\label{sec:hessian_properties}
The Hessian is a symmetric matrix $\mathbf{H}=\mathbf{H}^\top$ where each sub-block transforms under rotation as a Cartesian tensor
\begin{align}
    \mathbf{H}_{I,J} \xrightarrow[]{\mathbf{Q}} \mathbf{Q} \mathbf{H}_{I,J} \mathbf{Q}^\top 
    % \label{eq:hessian_symmetry}
\end{align}
or equivalently
\begin{align}
    \mathbf{H} \xrightarrow[]{\mathbf{Q}} (\mathbf{I}_{N} \otimes \mathbf{Q})\mathbf{H} (\mathbf{I}_{N} \otimes \mathbf{Q})^\top 
\end{align}
The symmetry follows directly from Schwarz's theorem, which states that for every scalar function $E(\mathbf{R})$ that has continuous partial second derivatives in the neighborhood of a point $\mathbf{R_0}$, the partial second derivatives commute
\begin{align}
    \left(\frac{\partial^2 E}{\partial \mathbf{R}_i \partial \mathbf{R}_j}\right)_{\mathbf{R}=\mathbf{R_0}} = \left(\frac{\partial^2 E}{\partial \mathbf{R}_j \partial \mathbf{R}_i}\right)_{\mathbf{R}=\mathbf{R_0}}
\end{align}
The transformation law under rotation follows straightforwardly from the chain rule: 
First, write the rotated coordinates as
\begin{align}
    \mathbf{R}' &= (\mathbf{I}_{N} \otimes \mathbf{Q}) \mathbf{R} 
    \frac{\partial \mathbf{R}'}{\partial \mathbf{R}} &= (\mathbf{I}_{N} \otimes \mathbf{Q})
\end{align}
Then, define the energy in the rotated frame
\begin{align}
    E'(\mathbf{R}') = E((\mathbf{I}_{N} \otimes \mathbf{Q})^\top \mathbf{R}') = E(\mathbf{R})
\end{align}
Then the first order derivative is
\begin{align}
    \nabla_\mathbf{R'}E'(\mathbf{R'}) &= (\mathbf{I}_{N} \otimes \mathbf{Q}) \nabla_\mathbf{R} E'(\mathbf{R}') \\
    &= (\mathbf{I}_{N} \otimes \mathbf{Q}) \nabla_\mathbf{R} E((\mathbf{I}_{N} \otimes \mathbf{Q})^{-1} (\mathbf{I}_{N} \otimes \mathbf{Q}) \mathbf{R}') \\
    &= (\mathbf{I}_{N} \otimes \mathbf{Q}) \nabla_\mathbf{R} E(\mathbf{R})
\end{align}
showing the equivariance of the forces. The second-order derivative is
\begin{align}
    \mathbf{H}' = \nabla^2_\mathbf{R'} E'(\mathbf{R'}) &= \nabla_\mathbf{R'} ((\mathbf{I}_{N} \otimes \mathbf{Q})\nabla_\mathbf{R}E(\mathbf{R})) \\
    &=(\mathbf{I}_{N} \otimes \mathbf{Q})\nabla_\mathbf{R'}\nabla_\mathbf{R}E(\mathbf{R}) \\
    &=(\mathbf{I}_{N} \otimes \mathbf{Q})\nabla_\mathbf{R}\nabla_\mathbf{R}E(\mathbf{R}) (\mathbf{I}_{N} \otimes \mathbf{Q})^\top \\
    &=(\mathbf{I}_{N} \otimes \mathbf{Q})\mathbf{H} (\mathbf{I}_{N} \otimes \mathbf{Q})^\top
\end{align}
which shows the Hessian's equivariance. 

\subsubsection{Non-integrability of Hessians\label{sec:conservative}}
% \subsection{Our Approach: Direct prediction of Hessians}
% We propose to directly predict Hessians with an equivariant neural network, instead of relying on automatic differentiation of an MLIP model. Similar ideas have been proposed for direct force predictions before \citep{gasteiger2021, liao2023equiformerv2}. While direct force predictions achieved low errors on benchmarks, they fell out of popularity for two reasons: 1) The main downstream application for MLIP forces is in MD simulations. Directly predicting forces means that the vector field is not guaranteed to be conservative, though, which causes MD-simulations to dissipate energy over time, leading to wrong observables. 2) The speed-up of direct force-prediction is rather modest. Backpropagation takes "only" about 2x-3x the time of a forward pass. 
% We argue that the situation is different for Hessians: 1) None of the most important use cases of Hessians rely on lengthy simulations where errors could accumulate. Instead, we usually only need Hessians at single points or for short optimizations. We will demonstrate that our directly predicted Hessians not only yield low errors on benchmarks but also achieve excellent performance on the most popular downstream applications. 2) Second-order automatic differentiation is much more expensive than first-order, as we will discuss below. Therefore, the savings for direct Hessian predictions are much larger.
Instead of relying on automatic differentiation, we propose to directly predict Hessians with an equivariant neural network. Similarly, direct force prediction has been proposed before \citep{gasteiger2021, liao2023equiformerv2}. 
While direct force prediction introduces non-integrable errors (non-conservative fields), these errors only cause unbounded blowups when integrated over time, as in Molecular Dynamics (MD). In contrast, for geometry optimization and standard Hessian applications (ZPE, transition state search, preconditioning), these errors remain bounded and merely limit convergence precision rather than accumulating. Therefore, direct Hessian predictions are safe for mainstream computational chemistry tasks, analogous to widely used but non-integrable methods like BFGS.

Since direct Hessian predictions are not formulated as derivatives, they are not guaranteed to be integrable. It is important to understand when a non-integrability error can lead to catastrophically wrong downstream predictions similar to direct force predictions in MD simulations. 
Let's first review the more familiar case: Why do direct force predictions fail in MD simulations but succeed in geometry optimization? Suppose we directly predict forces with error $e(x)=F_\text{pred}(x) - F(x)$. Write this error as an integrable part plus a remainder that is not a gradient,
\begin{align}
    e(x)=\nabla\phi(x)+S(x),
\end{align}
where \(\phi\) is a scalar potential and \(S(x)\in\mathbb{R}^{3N}\) is not the gradient of any scalar. Equivalently, \(\oint S\cdot\mathrm{d}x\neq 0\) for some closed path. 
Now consider how the energy of the system changes over time in an MD simulation:
$$
\begin{aligned}
\Delta E' &= \int_0^T F_\text{pred}(x(t))\cdot v(t)\, dt = \int_0^T \bigl(F(x(t)) + e(x(t))\bigr)\cdot v(t)\, dt \\
&= \int_0^T \bigl(F(x) + \nabla \phi(x) + S(x)\bigr)\cdot v(t)\, dt \\ 
&= E(T) - E(0) + \underbrace{\phi(T) - \phi(0)}_{\text{ error bounded with time}} + \underbrace{\int_0^T S(x)\cdot v(t)\, dt}_{\text{error unbounded with time}}
\end{aligned}
$$
Since MD simulations use large $T$, non-conservative forces lead to the blowups in error. 

In geometry optimization we define $E' = E + \phi$, then follow the gradient descent ODE:
\begin{align}
\frac{dE'(x(t))}{dt} &= \nabla E'(x(t)) \cdot \frac{d x(t)}{dt} \\
&= -\nabla E'(x(t)) \cdot (\nabla E(x) + e(x)) \\
&= -\nabla E'(x(t)) \cdot (\nabla E(x) + \nabla\phi(x) + S(x)) \\
&= -\nabla E'(x(t)) \cdot (\nabla E'(x)  + S(x)) \\
&= -|\nabla E'(x(t))|^2 + \nabla E'(x) \cdot S(x)
\end{align}
Using the general inequality $2ab \leq a^2 + b^2$:
\begin{align}
\frac{dE'(x(t))}{dt} &= -|\nabla E'(x(t))|^2 + \nabla E'(x) \cdot S(x) \\
&\leq -|\nabla E'(x(t))|^2 + \frac{1}{2}|\nabla E'(x(t))|^2 + \frac{1}{2}|S(x)|^2 \\
&=-\frac{1}{2}|\nabla E'(x(t))|^2 + \frac{1}{2}|S(x)|^2
\end{align}
Therefore, as long as $|\nabla E'(x(t))|^2 > |S(x)|^2$, the direct predicted force minimizes the energy plus some error potential similar to AD forces. Once we are close to convergence, we have $|\nabla E'(x(t))|^2 \leq |S(x)|^2$ and stop converging or enter a limit cycle. This means, we will converge to an extremum on an energy surface $E'$ within a distance of $O(|S|)$. In practice, we are interested in how close we converge to the extrema of the DFT energy $E$. For this, it does not matter if the error stems from the error potential $\phi$ shifting the extremum, or $S$ inducing a small limit cycle. 
As there is no time-dependent error blow-up, it is therefore safe to use direct forces for geometry optimization. 

Now consider a directly predicted Hessian \(H_\text{pred}(x)=H(x)+e(x)\). Split the matrix-valued error analogously into an integrable part and a remainder,
\begin{align}
    e(x)=\nabla^2\phi(x)+S(x),
\end{align}
where \(\nabla^2\phi\) is the Hessian of a scalar and \(S(x)\) is a symmetric matrix field that is not the Hessian of any scalar, i.e.\ it need not satisfy the mixed-partial compatibility conditions \(\partial_k S_{ij}=\partial_j S_{ik}\). 
A predicted Hessian is integrable if it is the Jacobian of a force field. Reconstructing the force along a trajectory then gives the path-independent identity \(\mathbf{F}(x(T))=\mathbf{F}(x(0))+\int_0^T H(x(t))\,\dot x(t)\,dt\). Substituting \(H_\text{pred}=H+\nabla^2\phi+S\) yields
\begin{align}
\mathbf{F}_{\mathrm{rec}}(T)
&=\mathbf{F}(x(0))+\int_0^T H_\text{pred}(x(t))\,\dot x(t)\,dt \\
&=\mathbf{F}(x(T))+\nabla\phi(T)-\nabla\phi(0)+\int_0^T S(x(t))\,\dot x(t)\,dt.
\end{align}
The last term is path-dependent and can grow with path length if \(S\dot x\) does not average to zero.
% Integrating that reconstructed force to recover the energy inherits the same path-dependent drift. 

None of the mainstream Hessian tasks reconstruct \(\mathbf{F}\) or \(E\) this way. ZPE and geometry classification are single-point evaluations of \(H\). The integrable error \(\nabla^2\phi\) and the remainder \(S\) therefore enter as local curvature errors of the same order. Mode following during transition-state search is equivalent to the geometry-optimization analysis with direct forces above. Energy minimization with a Hessian preconditioner is likewise local, in that the preconditioner cannot change the minimizer, only the convergence speed. BFGS is consistent with this picture. It maintains a local Hessian approximation, and is routinely used for energy minimization and transition-state search.

\subsection{Batched AD Hessian performance}
\label{sec:batched_ad_explanation}
\begin{figure}
    \centering
    \includegraphics[width=0.5\linewidth]{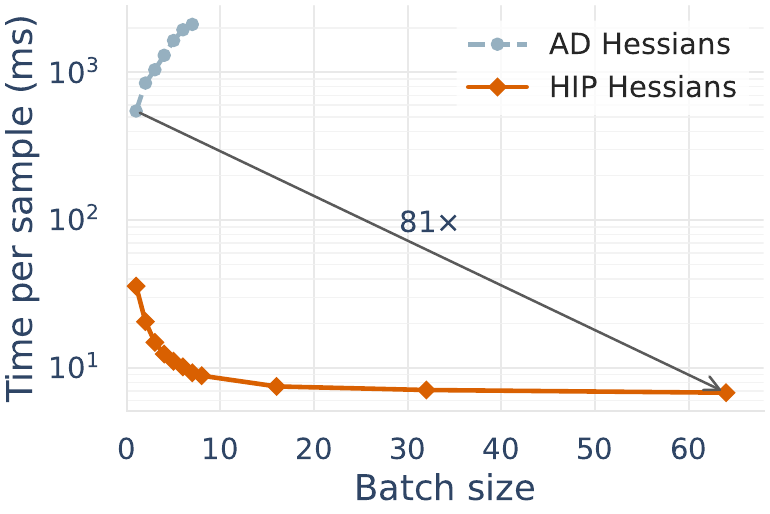}
    \caption{\textbf{Batch-wise Hessian timing.} Mean wall-clock time per sample versus batch size for HIP Hessians (orange) and AD Hessians (blue). AD timings stop at batch size 7 because larger batches exceeded available memory. HIP remains parallelizable to batch size 64. $n=10$ repeats per method and batch size.}
    \label{fig:batch_timing}
\end{figure}

To test the performance of batched Hessians, we measure the inference time per processed sample for increasingly larger batch sizes and plot the results in \ref{fig:batch_timing}.
We observe rapid degradation in the AD Hessian speed with increasing batch size. To explain this, we have to understand exactly how AD treats Hessians. From the point of view of the AD engine, there is no difference between a batch of two molecules of system size $N_A$ and $N_B$, and a larger system of size $N_A + N_B$. In both cases, the MLIP function takes in an array of dimension $3N_A + 3N_B$, and returns either a single scalar $E_A + E_B$ or even worse, an array $[E_A, E_B]$, so mathematically the energy function looks like $E(\mathbf{R}): \mathbb{R}^{3N_A + 3N_B} \rightarrow \mathbb{R}$ or $E(\mathbf{R}): \mathbb{R}^{3N_A + 3N_B} \rightarrow \mathbb{R}^2$. Consequently, a Hessian of this function is a block diagonal matrix of dimension $3(N_A + N_B) \times 3(N_A + N_B)$, or even $2\times 3(N_A + N_B) \times 3(N_A + N_B)$. Therefore, the memory costs grow quadratically with the batch size. Since we have to implement the Hessian calculation using sequential Hessian-Vector products to avoid running out of memory, we cannot even take advantage of parallelization in batched Hessian processing.

\subsection{Additional results}
\begin{figure}
    \centering
    \includegraphics[width=0.99\linewidth]{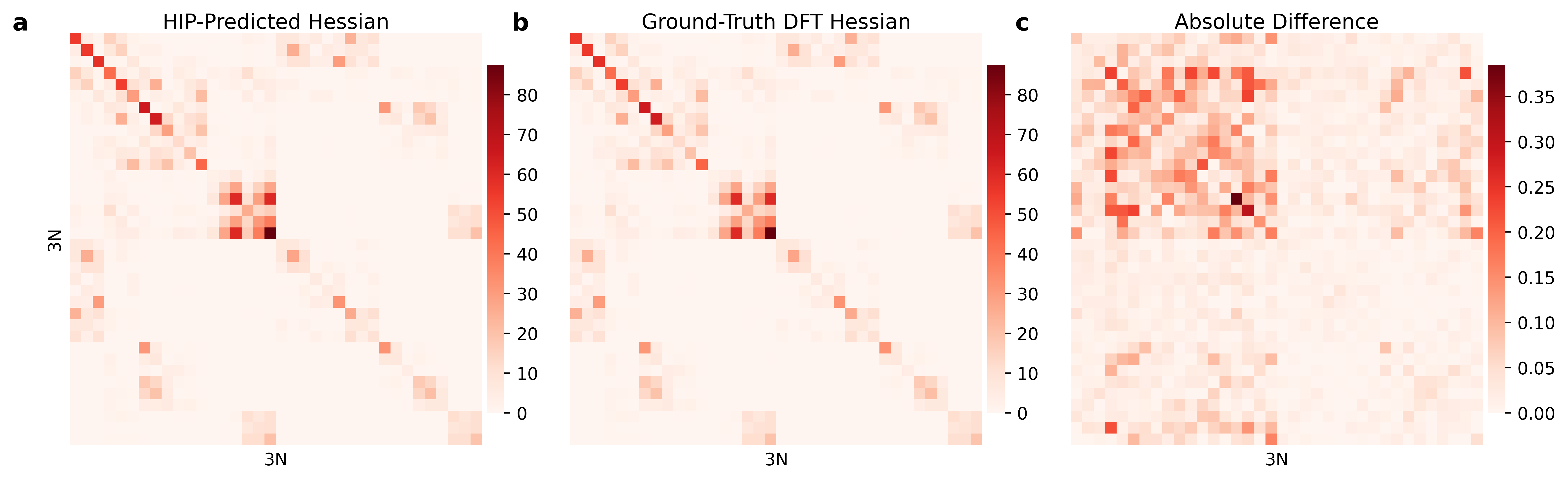}
    \caption{\textbf{HIP compared to the ground-truth DFT Hessian.} Absolute values of Hessian entries for test sample C$_5$H$_6$O, in eV/\AA$^2$. Axes index the $3N\times 3N$ matrix elements.}
    \label{fig:hessian_heatmap}
\end{figure}

\subsubsection{Loss function\label{sec:lossablation}}
We are trying to design a loss function that emphasizes the lowest lying eigenvalues/eigenvectors.
To design such a loss function, one could naively compute a loss directly as $L= \sum_i^k |\mathbf{v}_i^\text{predict} - \mathbf{v}_1^\text{true}|$. This requires computing the eigenvalues and eigenvectors of the true and predicted Hessian via an eigenspectrum decomposition. Unfortunately, backpropagation through such an eigenspectrum decomposition is numerically unstable \citep{aaComputationEigenvalueEigenvector2007}.
First, gradients computed using the eigenvectors tensor will only be finite when $\mathbf{H}$ has distinct eigenvalues. Furthermore, if the distance between any two eigenvalues is close to zero, the gradient will be highly sensitive, as it depends on the eigenvalues $\lambda_i $ through the computation of
$
\min_{i \neq j} \frac{1}{\lambda_i - \lambda_j}
$.
Instead, we propose the following subspace loss.

In the following section, we compare the eigenspace loss that we introduced in \eqref{eq:eigenloss} to the standard choice of mean square error (MSE) and mean absolute error (MAE). 

In particular, we emphasize the subspace $k=8$. This is because transition state search, ZPE, and frequency analysis for extrema classification all depend on the two lowest-lying eigenvectors and eigenvalues. We choose $k=8$ instead of $k=2$ to account for the usual five or six redundant translational and rotational degrees of freedom, which we do not remove during training for numerical stability.

We combine the subspace loss with MAE, so the total loss becomes
\begin{align}
    \mathcal{L}_\text{MAE+Sub} 
    &= \mathcal{L}_\text{MAE} + \mathcal{L}_\text{sub}^{(k=8)} \\
    &=
    \sum_{i,j} |\mathbf{H}_{i,j} - \mathbf{H}^\text{pred}_{i,j}|
     + \sum_{i,j} \left|\mathbf{V}_{[:,:k]}^\top \mathbf{H}^\text{pred}\mathbf{V}_{[:,:8]} - \mathbf{\Lambda}_{[:,:8]}\right|_{i,j} 
     % + \gamma \sum_{i,j} \left|\mathbf{V}_{[:,k:]}^\top \mathbf{H}^\text{pred}\mathbf{V}_{[:,k:]} - \mathbf{\Lambda}_{[:,k:]}\right|_{i,j}
\end{align}
For simplicity, we weight both the MAE and subspace loss equally. Future work could benefit from carefully tuning the relative weights between the loss terms.

\begin{figure}
    \centering
    \includegraphics[width=0.75\linewidth]{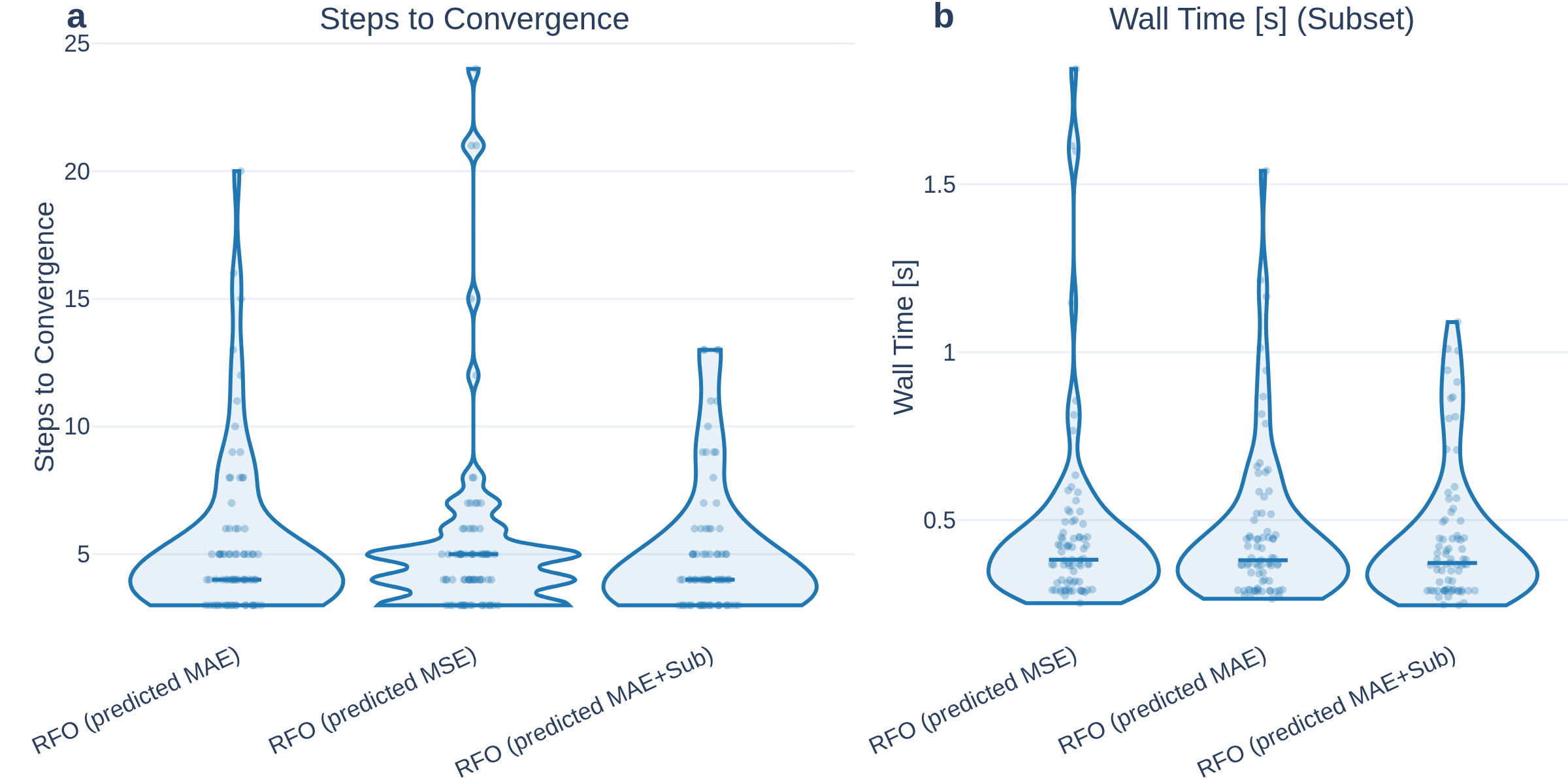}
    \caption{\textbf{Geometry relaxation by loss function.} RFO with HIP-EquiformerV2 Hessians trained using MSE, MAE or MAE+Sub. \textbf{(a)} Steps to convergence. \textbf{(b)} Wall-clock time. Violins show the sample distribution and points are individual geometries. $n=80$ Transition1x validation reactant geometries.}
    \label{fig:loss_relax}
\end{figure}

On downstream tasks, subspace loss improves over MAE and MSE in (i) transition state search, as shown in Table \ref{tab:loss_tssearch}, and (ii) transition state identification frequency analysis, as shown in Table \ref{tab:loss_freqanalysis}. We observe equal performance between MAE and the subspace loss for (i) geometry relaxation Table \ref{fig:loss_relax} and (ii) computing the zero-point energy Table \ref{tab:loss_zpe}.

Table \ref{tab:loss_acc} shows that the subspace loss improves the first eigenvector $\bm{v}_1$ cosine similarity, and the second eigenvalue $\lambda_2$ MAE. This is consistent with the better performance of the subspace loss on transition state-related tasks, since the transition state is characterized via the two smallest eigenvalues $\lambda_1, \lambda_2$, and is found by following the first eigenvector $\bm{v}_1$.

MSE generally underperforms compared to MAE and MAE with subspace loss.
MSE has the theoretical advantage of being rotation-invariant, whereas MAE is not.
In practice, however, we find that the MSE leads to a high variance (spikes) in the training loss and gradient norm across batches. We speculate that MAE outperforms MSE due to training stability.

Hessian prediction with any loss function tested significantly outperforms prior AD Hessians.

% Accuracy
\begin{table}
\centering
\resizebox{0.99\textwidth}{!}{
\centering
\begin{tabular}{lccccccccc}
\hline
\multirow{2}{*}{Loss}  & Hessian $\downarrow$  & Eigenvalues $\downarrow$ & CosSim $\bm{v}_1$ $\uparrow$ & $\lambda_1$ $\downarrow$ & $\lambda_2$ $\downarrow$ & $\lambda_1^{E}$ $\downarrow$ & $\lambda_2^{E}$ $\downarrow$ & CosSim $\bm{v}_1^{E}$ $\uparrow$ & CosSim $\bm{v}_2^{E}$ $\uparrow$ \\
 & eV/\AA$^2$ & eV/\AA$^2$ & unitless & eV/\AA$^2$ & eV/\AA$^2$ & eV/\AA$^2$ & eV/\AA$^2$ & unitless & unitless \\
\hline
MSE  & 0.033 & 0.072 & 0.728 & 0.168 & 0.088 & 0.037 & 0.013 & 0.959 & 0.928\\
MAE  & \textbf{ 0.026 } & \textbf{ 0.057 } & 0.825 & \textbf{ 0.127 } & 0.052 & \textbf{ 0.031 } & 0.012 & 0.976 & 0.952 \\
MAE+Sub  & 0.030 & 0.063 & \textbf{ 0.870 } & 0.130 & \textbf{ 0.030 } & 0.031 & \textbf{ 0.010 } & \textbf{ 0.980 } & \textbf{ 0.957 } \\
\hline
\end{tabular}
}
\caption{\textbf{Loss-function accuracy.} Mean absolute errors and cosine similarities for HIP-EquiformerV2 trained with MSE, MAE or MAE plus subspace loss (MAE+Sub). Hessian, eigenvalue and $\lambda_1$, $\lambda_2$ columns are in eV/\AA$^2$. Superscript $E$ denotes quantities after Eckart projection of the mass-weighted Hessian. Bold values mark the best result in each column. $n=1000$ HORM-T1x validation geometries.}
\label{tab:loss_acc}
\end{table}

% Second order optimization

\begin{table}
\centering
\resizebox{0.7\textwidth}{!}{
\centering
\begin{tabular}{llll}
\toprule
Hessian & Model & ZPE MAE (Std) [eV] & $\Delta$ZPE MAE (Std) [eV] \\
\midrule
MSE & HIP-EquiformerV2 & 0.0008 (0.0012) & 0.0019 (0.0025) \\
MAE & HIP-EquiformerV2 & 0.0004 (0.0004) & \textbf{0.0013 (0.0019)} \\
MAE+Sub & HIP-EquiformerV2 & \textbf{0.0004 (0.0003)} & 0.0016 (0.0018) \\
\bottomrule
\end{tabular}
}
\caption{\textbf{Zero-point energy by loss function.} MAE of the reactant ZPE and of the reaction $\Delta$ZPE relative to DFT for HIP-EquiformerV2 trained with MSE, MAE or MAE+Sub. Values in parentheses are sample standard deviations of the signed errors. Bold values mark the best result in each column. Energies are in eV per molecule. $n=47$ reactant--product pairs.}
\label{tab:loss_zpe}
\end{table}

% TS search
\begin{table}
\centering
\resizebox{0.6\textwidth}{!}{
\centering
\begin{tabular}{llcccc}
\hline
Loss & Model & TS Success & RFO Converged & Both \\
\hline
MSE & HIP-EquiformerV2 & 718 & 669 & 659 \\
MAE & HIP-EquiformerV2 & 740 & 693 & 683 \\
MAE+Sub & HIP-EquiformerV2 & \textbf{750} & \textbf{705} & \textbf{698} \\
\hline
\end{tabular}
}
\caption{\textbf{Transition-state search by loss function.} Counts of ReactBench reactions that pass TS success, RFO convergence, or both, for HIP-EquiformerV2 trained with MSE, MAE or MAE+Sub. Bold values mark the best result in each column. $n=960$ reactions.}
\label{tab:loss_tssearch}
\end{table}

\begin{table}
\centering
\resizebox{0.75\textwidth}{!}{
\centering
\begin{tabular}{lccccccc}
\hline
Name & TPR $\uparrow$ & FPR $\downarrow$ & FNR $\downarrow$ & TNR $\uparrow$ & Precision $\uparrow$ & Accuracy $\uparrow$&  Accuracy All $\uparrow$ \\
\hline
MSE  & 88\% & 11\%  & 12\%  & 89\% & 86\% & 89\% & 86\% \\
MAE  & 93\% & 7\%  & 7\%  & 93\% & 91\% & 93\% & 91\%  \\
MAE+Sub  & \textbf{93\%} & \textbf{6\%}  & \textbf{7\%}  & \textbf{94\%} & \textbf{92\%} & \textbf{94\%} & \textbf{92\%} \\
\hline
\end{tabular}
}
\caption{\textbf{Frequency-analysis classification by loss function.} True-positive rate (TPR), false-positive rate (FPR), false-negative rate (FNR), true-negative rate (TNR), precision and accuracy for identifying first-order transition states from the number of negative Hessian eigenvalues. Accuracy All is the accuracy for classifying any stationary-point type (minimum, first-order saddle or higher-order saddle). HIP-EquiformerV2 was trained with MSE, MAE or MAE+Sub. Bold values mark the best result in each column. $n=1000$ HORM-T1x validation geometries.}
\label{tab:loss_freqanalysis}
\end{table}

% \subsubsection{Computational cost compared to finite difference}
% \begin{figure}
%     \centering
%     \includegraphics[width=0.8\linewidth]{plots/speed_memory_finitedifference.png}
%     \caption{
%     \textbf{Computational cost.} Comparing (a) time and (b) memory of a forward without Hessians, with HIP, automatic differentiation (AD), finite difference Hessians. The memory for Forward Pass, FD Hessians, and HIP Hessians is nearly identical.
%     }
%     \label{fig:speedmemoryfinitedifference}
% \end{figure}
% \cite{wander2025accessing} showed that finite-difference Hessians obtained from finetuned MLIPs can increase success rates in transition state search for catalysis. The advantage of finite difference is that it does not require additional memory over a regular forward pass, and only force labels are needed for training. The limitation of finite-difference Hessians is their high computational cost. Like AD Hessians, finite difference requires $O(N^2)$ time, but with an even larger prefactor, as can be seen in Fig. \ref{fig:speedmemoryfinitedifference}.

\subsection{Experimental Details}

\subsubsection{Training}

The hyperparameters of our HIP-EquiformerV2 model are listed in \ref{tab:hyperparameters}. We inherit the model settings and large parts of the optimizer settings from the HORM codebase \citep{cui2025horm}.

\begin{table}[h]
\centering
\resizebox{0.55\textwidth}{!}{ % Resize to fit the page width
\begin{tabular}{ll}
\hline
% \textbf{Category} & \textbf{Hyperparameter (Value)} 
\hline
Model & Type: EquiformerV2 \\
      & Layers: 4 \\
      & Sphere channels: 128 \\
      & Attention hidden channels: 64 \\
      & Attention heads: 4 \\
      & Attention alpha channels: 64 \\
      & Attention value channels: 64 \\
      & FFN hidden channels: 128 \\
      & Activation: SiLU \\
      & Distance basis: 512 (Gaussian) \\
      & Radius cutoff: 12 Angstrom \\
      & Cutoff Hessian: 12 Angstrom \\
      & Hessian layers: 3 (1 for end-to-end) \\
      & Spherical harmonics: Lmax=4, Mmax=2 \\
      & Grid resolution: 18, Sphere samples: 128 \\
      & Dropout: $\alpha$=0, drop path=0 \\
\hline
Loss & MAE weight: 1 \\
     & Eigen subspace $k$: 8 \\
     & Eigen loss weight $\alpha$: 1.0 \\
\hline
Optimizer & AdamW \\
          & Betas: (0.9, 0.999) \\
          & AMSGrad: True \\
          & Weight decay: 0 \\
        & Batch size: 128 \\
         & Gradient clipping: 0.1 (norm) \\
\hline
Learning Rate 
          & Warmup: 15k steps, 0 to $5\times10^{-4}$ \\
        & Scheduler: cosine decay over 500k steps \\
        & Final learning rate: $5\times10^{-5}$ \\
Trainer & Steps: 500k (2-3 days on one H100 GPU) \\
\hline
\hline
\end{tabular}
} % end resizebox
\caption{\textbf{Hyperparameters}. For simplicity, we use the same hyperparameters for HIP as HORM \citep{cui2025horm}.
}
\label{tab:hyperparameters}
\end{table}

% \paragraph{Training resources\label{sec:trainingresources}}
Training AD Hessians is exceptionally more expensive than training HIP.
The memory cost of computing loss gradients through the full  $3N\times3N$ Hessian is prohibitive. Instead, the baseline models from \cite{cui2025horm} randomly sample one or two columns of the Hessian via random Hessian-vector products at each training step. Despite this, training AD Hessians remains exceedingly expensive. The HORM paper states that they use 2 HVPs at a batch size of 128. Their checkpoint metadata shows a per-device batch size of 8, suggesting a 16$\times$ A30 GPU setup for 1 million gradient steps.
In contrast, we trained HIP on a single H100 GPU for 500k steps. Additionally, each HIP training step costs 1/5 as much as AD Equiformer training on our hardware. Finally, we observe that HIP fully converges after 300-500k training steps, whereas AD Hessians require multiple steps due to the limited supervision and the inability to use the full Hessian.

\begin{table}[h]
\centering
\renewcommand{\arraystretch}{1.6} % Adjust row height
\resizebox{0.8\textwidth}{!}{ % Resize to fit the page width
\begin{tabular}{lccccccc}
\hline
\textbf{Model} & Hessian Columns & Batch Size &
BZ / GPU & GPU & \# GPUs & Training Steps \\
\hline
AlphaNet            & 1 & 32  & 16 & H20 & 2  & 688,000 \\
LeftNet-CF          & 1 & 64  & 16 & A30 & 8  & 192,000 \\
LeftNet-DF          & 2 & 64  & 16 & A30 & 8  & 120,000 \\
EquiformerV2 (E-F)  & 0 & 128 & 8  & A30 & 16 & 384,000 \\
EquiformerV2        & 2 & 128 & 8  & A30 & 16 & 1,008,000 \\
\hline
HIP-EquiformerV2        & 3N (all) & 128 & 128 & H100 & 1 & 500,000 \\
\hline
\end{tabular}
}
\caption{\textbf{Training resources.} Hardware, Hessian-column sampling, batch size and training-step counts for HIP-EquiformerV2 compared with HORM checkpoints \citep{cui2025horm}. Hessian Columns is the number of Hessian-vector products used per training step; HIP uses the full $3N\times 3N$ Hessian.}
\label{tab:traingpu}
\end{table}

\subsubsection{Geometry Optimization \label{sec:relax_methods}}
We compare our RFO-based optimization \citep{RSRFO1998a} using predicted Hessians (RFO predicted), and our RFO with BFGS updates with initial predicted Hessian (RFO-BFGS predicted init) against first-order methods, quasi-second-order BFGS variants, and exact second-order approaches obtained via automatic differentiation or finite differences. 

Our baselines include steepest descent with backtracking line search (SteepestDescent) and the first-order method FIRE \citep{bitzek2006}. For quasi-second-order methods, we consider RFO with BFGS updates initialized from the identity (RFO-BFGS). For full second-order methods, we include RFO with either automatic differentiation or finite difference Hessians. 
%Namely we report
%(a) steepest descent with backtracking line search (SteepestDescent)
%(b) fast inertial relaxation engine (FIRE) 
%(c) RFO with identity matrix as initial guess, followed by BFGS (RFO-BFGS unit-init)
%(e) RFO+BFGS with learned Hessian as initial guess (RFO-BFGS predicted-init) (ours)
% (f) RFO+BFGS where the Hessian is predicted every third step (RFO-BFGS Hpred-3)
%(f) RFO with predicted Hessian at every step (RFO predicted) (ours)
%(g) RFO+BFGS with AD Hessian as initial guess (RFO-BFGS AD-init)
%(h) RFO with AD Hessian at every step (RFO AD)
%
We run relaxations on 80 reactant geometries from the Transition1x validation set (which is different from HORM-T1x), that we noise with $0.5$ \AA{} RMS. We use the RS-RFO and redundant coordinate implementation in pysisyphus \citep{steinmetzer2021, herbolComputationalImplementationNudged2017}. We set the convergence criteria to "Gaussian default" (see \ref{sec:convergence}), with a budget of 150 steps.

\subsubsection{Transition State Search with ReactBench\label{sec:reactbench}}
We use the recently introduced \href{https://github.com/deepprinciple/ReactBench}{ReactBench} benchmark \citep{zhao2025}, which relies on pysisyphus \citep{steinmetzer2021} and PyGSM\cite{aldazZimmermanGroupPyGSM2025} for implementation of the optimization routines.
The workflow consists of generating an initial guess using the growing string method \citep{aldazZimmermanGroupPyGSM2025, zimmerman2015}, followed by local search using the restricted-step partitioned rational-function-optimization method (RS-P-RFO) \citep{RSRFO1998a}. Finally, convergence to the correct transition state is confirmed by frequency analysis and following the intrinsic reaction coordinate (IRC).
% minimum energy reaction pathway (MERP) in mass-weighted cartesian coordinates between the transition state of a reaction and its reactants and products
Following previous work \citep{zhao2025, cui2025horm}, we report success metrics of each step: 
\begin{enumerate}
    \item 
GSM successfully converged below a force RMS of $5e^{-5}$ Hartree/Bohr within 100 iterations ("GSM Success")
\item After RS-P-RFO, the frequency analysis determines geometry as a true transition state ("TS Success")
\item IRC converged to the same criteria and yields the original initial reactant and product ("IRC Verified").
\end{enumerate}
We then treat the samples successfully passing (a)-(c) as transition state proposals, and verify for a random subset of 100, if the DFT Hessians have one negative eigenvalue and the DFT force RMS is below $2e^{-3}$ Ha/Bohr. The verification single points use GPU4PySCF v1.3.0 \citep{wu2025gpu4pyscf} at $\omega$B97X/6-31G*. We additionally report if RS-P-RFO converged to "Gaussian default" (\ref{sec:convergence}) within 50 steps.

\subsubsection{Zero-Point Energy\label{sec:zpe_setup}}
Starting from 80 reactant and product geometries from the Transition1x validation set, we relax the geometries with the RS-RFO implementation in pysisyphus \citep{steinmetzer2021}, using BFGS updates and identity Hessian initialization, on the MLIP trained at $\omega$B97X/6-31G*.
We use Gaussian tight convergence (see \ref{sec:convergence}).
We compute ZPE from the Eckart-projected, mass-weighted Hessian, \(\tilde{H}_{ij}=H_{ij}/\sqrt{m_i m_j}\), and report energies in eV per molecule.
Absolute ZPE error is evaluated at the reactant geometry.
For method $m$,
$$
\Delta \mathrm{ZPE}_{m}
= \mathrm{ZPE}_{m}(\text{product}) - \mathrm{ZPE}_{m}(\text{reactant}),
$$
and the model error is
$$
\bigl|\Delta \mathrm{ZPE}_{\text{model}} - \Delta \mathrm{ZPE}_{\text{DFT}}\bigr|,
$$
where the DFT reference is the $\omega$B97X/6-31G* Hessian at the same (MLIP-relaxed) geometry. 

\subsubsection{Convergence Criteria\label{sec:convergence}}
Throughout our experiments, we adopt the following convergence criteria, as shown in table \ref{tab:convergence_criteria}, set forth by the Gaussian software and widely used across different codebases.
All criteria need to be met for a geometry to be considered converged.

\begin{table}
\centering
\resizebox{0.75\textwidth}{!}{
\centering
\begin{tabular}{lccccc}
\toprule
Setting & Max Force & RMS Force & Max Step & RMS Step & Used for \\
\midrule
% Gaussian loose             & 1.7e-3 & 1.0e-2 & 6.7e-3 & &  \\
Gaussian default           & 4.5e-4 & 3.0e-4 & 1.8e-3 & 1.2e-3 & Relaxation, TS search \ref{sec:relax} \\
Gaussian tight             & 1.5e-5 & 1.0e-5  & 6.0e-5 & 4.0e-5 & ZPE \\
% Gaussian very tight        & 1.0e-6 &         & 6.0e-6 & 4.0e-6 & 
\bottomrule
\end{tabular}
}
\caption{Convergence criteria used in our experiments}
\label{tab:convergence_criteria}
\end{table}

\subsubsection{Glycine proton transfer\label{sec:glycinedetails}}

To assess whether the predicted Hessians encode physically meaningful curvature near a transition state, we selected a reaction for which the relevant chemical motion can be described by simple internal coordinates. We first annotated all non-training reactions in Transition1x by comparing reactant and product bond graphs, excluding the training split and considering both the validation and test splits. For each reaction, we recorded the bonds formed and broken, the number of connected molecular components, and a coarse reaction-family label. We then prioritized reactions with a small number of bond changes and an unambiguous two-dimensional representation of the reaction coordinate.

We chose the intramolecular proton transfer of glycine, present in the Transition1x test split (sample ID 5, rxn1961). Glycine is the simplest amino acid, and its neutral-zwitterionic tautomerization,
\(
\mathrm{NH_2CH_2COOH \rightleftharpoons NH_3^+CH_2COO^-},
\)
has been widely studied as a model proton-transfer process. In this reaction, the acidic carboxyl proton is transferred between an oxygen and the amine nitrogen, interconverting O--H- and N--H-bonded forms. The transferring proton is therefore naturally described by the two distances
\(
q_{\mathrm{NH}} = d(\mathrm{N4},\mathrm{H9}), \qquad
q_{\mathrm{OH}} = d(\mathrm{O3},\mathrm{H9}).
\)
Previous studies of glycine tautomerization have commonly used a proton-transfer coordinate based on these distances
\(\xi = q_{\mathrm{NH}} - q_{\mathrm{OH}}\) (see e.g. \cite{zhang2024glycine}). 

% \paragraph{2D collective variables}
% Here, we instead use a two-dimensional coordinate system consisting of the antisymmetric proton-transfer coordinate
% \[
% s = q_{\mathrm{NH}} - q_{\mathrm{OH}}
% \]
% and the symmetric compression coordinate
% \[
% \sigma = q_{\mathrm{NH}} + q_{\mathrm{OH}}.
% \]
% This representation separates progress along the proton-transfer coordinate from changes in the total N--H/O--H distance and provides a compact visualization of the local potential-energy surface around the transition state.

We first investigate a two-dimensional surface around the proton transfer.
For each point on the 
% \((\xi,\sigma)\) 
\((q_{\mathrm{NH}}, q_{\mathrm{OH}})\)
surface, we constructed an initial geometry by starting from the Transition1x \citep{schreiner2022transition1x} transition-state structure and placing the transferring proton to satisfy the target values of \(q_{\mathrm{NH}}\) and \(q_{\mathrm{OH}}\). We excluded any pair \((q_{\mathrm{NH}}, q_{\mathrm{OH}})\) that violated the triangle inequality
\(
q_{\mathrm{NH}} + q_{\mathrm{OH}} < d(\mathrm{N4},\mathrm{O3})
\quad \mathrm{or} \quad
|q_{\mathrm{NH}} - q_{\mathrm{OH}}| > d(\mathrm{N4},\mathrm{O3}).
\)
The remaining structures were relaxed while constraining \(q_{\mathrm{NH}}\) and \(q_{\mathrm{OH}}\), allowing all other degrees of freedom to relax. 
They were first relaxed with GFN2-xTB, using TBLite and ASE BFGS.
% from ase.constraints import FixInternals
% from ase.optimize import BFGS
% from tblite.ase import TBLite
% atoms.set_constraint(FixInternals(bonds=[[q_nh, [N_ATOM, H_ATOM]], [q_oh, [O_ATOM, H_ATOM]]]))
% atoms.calc = TBLite(method="GFN2-xTB", verbosity=0)
% opt = BFGS(atoms, logfile=None)
% opt.run(fmax=0.05, steps=300)
They were then reoptimized at the DFT level using ORCA 6.1.1 with
\texttt{! wB97X-D3 6-31G(d) TightSCF Opt}
and the same two bond-distance constraints. This produced 579 DFT-relaxed geometries.

\begin{figure}
    \centering
    \includegraphics[width=0.52\linewidth]{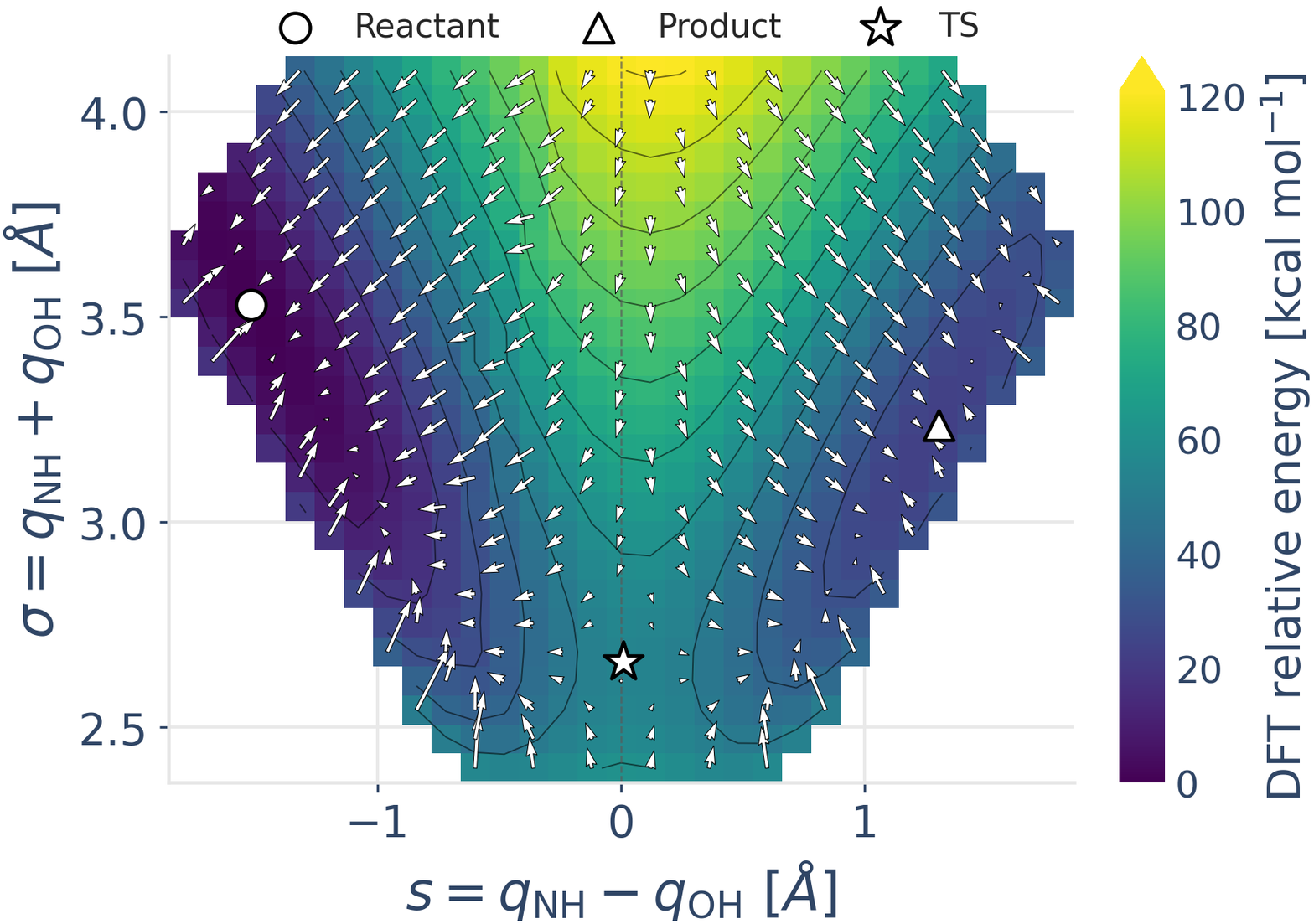}
    \includegraphics[width=0.45\linewidth]{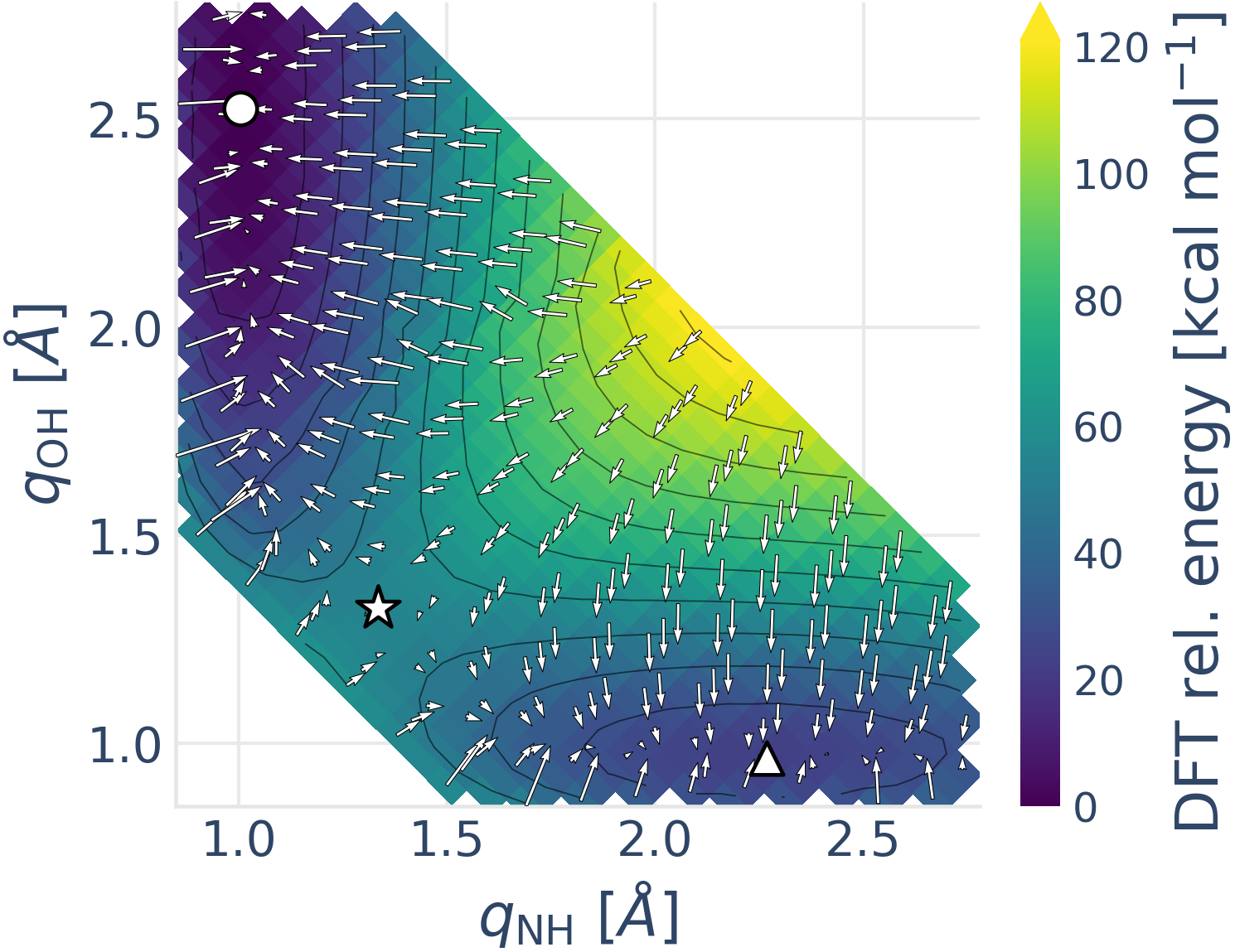}
     \caption{
     \textbf{Glycine proton-transfer energy surface.}
     DFT relative energy at $\omega$B97X-D3/6-31G(d) on the two-dimensional proton-transfer surface, $n=579$ constrained-relaxed geometries. Colour is DFT relative energy in kcal\,mol$^{-1}$. White arrows are forces in the plotted coordinates. The circle, star and triangle mark the DFT reactant, transition state and product. \textbf{(Left)} Coordinates $s=q_{\mathrm{NH}}-q_{\mathrm{OH}}$ and $\sigma=q_{\mathrm{NH}}+q_{\mathrm{OH}}$. \textbf{(Right)} Distances $q_{\mathrm{NH}}$ and $q_{\mathrm{OH}}$. The DFT reactant and product have no imaginary frequencies; the DFT transition-state geometry has one.
     }
     \label{fig:glycine_forcefield}
\end{figure}

For each DFT-relaxed geometry, we computed energies, forces, and Hessians using ORCA 6.1.1 \cite{ORCA} with the \(\omega\)B97X-D3 functional and the 6-31G(d) basis set. Geometry optimizations used
\texttt{! wB97X-D3 6-31G(d) TightSCF Opt}
with the two proton-transfer bond distances constrained. Energies and forces were then evaluated with
\texttt{! wB97X-D3 6-31G(d) TightSCF EnGrad},
and Hessians were evaluated with
\texttt{! wB97X-D3 6-31G(d) TightSCF Freq}.
The data is available at \url{huggingface.co/andreasburger/hip/tree/main/orca_wb97x_d3_631gd_glycine_pt_dft_relaxed_579}.
The resulting energy surface is shown in figure~\ref{fig:glycine_forcefield}.

% \paragraph{1D minimum energy path}
We also study the one-dimensional minimum energy path for the same glycine proton-transfer reaction to evaluate model behaviour along a chemically meaningful trajectory. We used the stored Transition1x record, which contains 266 geometries, which we interpreted as an initial 10-image NEB path followed by 32 saved blocks of 8 internal NEB/CINEB images. We used the final saved block of 8 with the reactant and product geometries as the 10-image minimum energy path.

To obtain a smooth trajectory for evaluation and visualization, we connected adjacent images along this path using geodesic interpolation in redundant internal-coordinate space. The original NEB images were kept fixed. This produced 145 path geometries. All methods were evaluated on the same fixed structures. Again, single-point energies, forces, and Cartesian Hessians were computed for each geometry using ORCA 6.1.1 with \texttt{! wB97X-D3 6-31G(d) TightSCF EnGrad Freq} and are available at \url{huggingface.co/andreasburger/hip/tree/main/orca_wb97x_631gd_glycine_mep_145}.

\subsubsection{Size generalization on PubChem\label{sec:pubchemorca}}

To test Hessian prediction beyond the HORM training distribution, we assembled a benchmark of 710 neutral organic molecules containing only C, H, N, and O, with 30-100 atoms (71 atom-count bins, 10 molecules per bin). Geometries were drawn from the PubChem database via the PUG REST API. For each target atom count \(N\), we enumerated stoichiometrically plausible neutral CHNO formulas and queried PubChem for matching compounds (`fastformula` endpoint, up to 12 CIDs per formula). Candidates were retained only if they satisfied 
(a) formal charge 0 (PubChem `Charge` property)
(b) composition restricted to \(\{Z=1,6,7,8\}\) (c) exactly \(N\) atoms in the 3D structure.

We first attempted to retrieve PubChem 3D conformers (SDF). When no suitable 3D structure was available, we generated a conformer from the isomeric SMILES using RDKit (ETKDGv3 embedding followed by MMFF or UFF relaxation). Uncovered atom-count bins were additionally filled from deterministic RDKit-generated CHNO conformers. 
Reference energies, Cartesian gradients, and Hessians were computed with ORCA~6
at $\omega$B97X/6-31G(d), \texttt{TightSCF}, on the fixed input geometries
(charge~$0$, multiplicity~$1$).
We used two single-point jobs per molecule:
(i)~a \texttt{Freq} calculation for the analytical Hessian, and
(ii)~an \texttt{EnGrad} calculation for the energy and gradient.
\begin{verbatim}
! wB97X 6-31G(d) TightSCF Freq
! wB97X 6-31G(d) TightSCF EnGrad
%pal nprocs 8 end
%maxcore 4000
\end{verbatim}
The reference dataset and geometries are released under \url{huggingface.co/andreasburger/hip/tree/main/orca_wb97x_631gd_chno_30_100}.

For aggregate PubChem plots, we remove pathological molecules before bin-wise averaging.
Outliers are identified from the per-molecule Cartesian Hessian MAE,
\(
  \mathrm{MAE}(\mathbf{H}) = \frac{1}{9N^2}\sum_{i,j=1}^{3N}\left|H^{\mathrm{pred}}_{ij}-H^{\mathrm{ref}}_{ij}\right|,
\)
where $N$ is the number of atoms.
For each atom-count bin $N$, let $\{x_i\}_{i=1}^{n_N}$ denote the Hessian MAEs of molecules in that bin, with $n_N=10$.
We compute the bin median $\tilde{x}_N$ and the median absolute deviation
\(
  \mathrm{MAD}_N = \operatorname{median}_i\!\left(\left|x_i-\tilde{x}_N\right|\right).
\)
The modified $z$-score of sample $i$ is
\(
  M_i = 0.6745\,\frac{x_i-\tilde{x}_N}{\mathrm{MAD}_N},
\)
following Iglewicz and Hoaglin \citep{iglewicz1993outliers}. This modified $z$-score uses the median instead of the mean, and MAD instead of the variance, to improve robustness. The prefactor $0.6745$ is the median absolute deviation of a standard normal variable.
Sample $i$ is flagged as an outlier if it exceeds $|M_i|>10$.
Outliers are detected independently for each method, and the union of flagged
samples is removed from all methods so that curves remain comparable.
This removes 10 of 710 PubChem molecules.
We apply the same procedure to HORM-RGD1, detecting outliers independently for HIP, AD, and EquiformerV2 trained on energies and forces only (E-F), and remove 72 of 60,000 molecules.

Because raw Cartesian Hessian eigenvalues mix internal and external (translation/rotation) modes, we additionally project both reference and predicted Hessians onto the vibrational subspace using Eckart projection (mass-weighted, translation-rotation removal) and compare the resulting eigenvalue spectra. These projected eigenvalue errors are used in Fig.~\ref{fig:speed_memory}c-d. The eigenvalue MAE is computed as
\(
\mathrm{MAE}_\lambda = \langle |\lambda^{\mathrm{pred}}_k - \lambda^{\mathrm{ref}}_k| \rangle
\)
where \(\{\lambda_k\}\) are the projected vibrational eigenvalues $\mathrm{eV}/\mathring{\mathrm{A}}^2$.
% We also report agreement on the number of imaginary modes (negative eigenvalues after projection).

\end{document}